\documentclass[10pt,twocolumn,letterpaper]{article}

\usepackage[pagenumbers]{wacv} 

\usepackage[table]{xcolor}

\newcommand{\fcirc}[1]{%
  \tikz[baseline=(C.base)]\node[%
    circle, fill=black, inner sep=0.6pt, minimum size=1.55ex
  ] (C) {\textcolor{white}{\scriptsize #1}};%
}

\usepackage{amsmath,amssymb,bm}
\usepackage{algorithm}
\usepackage{algpseudocode}
\algrenewcommand\algorithmicrequire{\textbf{Input:}}
\algrenewcommand\algorithmicensure{\textbf{Output:}}
\algnewcommand{\LineComment}[1]{\State \(\triangleright\) \textit{#1}}

\usepackage{tikz}
\usetikzlibrary{positioning,fit,arrows.meta,backgrounds,calc}
\usepackage{pgfplots}
\pgfplotsset{compat=1.18}

\usepackage{enumitem}

\usepackage{mwe}

\definecolor{wacvblue}{rgb}{0.21,0.49,0.74}
\usepackage[pagebackref,breaklinks,colorlinks,allcolors=wacvblue]{hyperref}

\def\wacvPaperID{1231} 
\def\confName{WACV}
\def\confYear{2027}

\title{TORINO: Token Reduction via Interpretable Concept Overlap in Vision-Language Models}

\author{
Riccardo Renzulli$^{1}$ \quad Gabriele Spadaro$^{1}$ \quad Shruthi Gowda$^{2}$\\[0.2em]
Alaa Eddine Mazouz$^{3}$ \quad Van-Tam Nguyen$^{3}$\\[0.4em]
$^{1}$University of Turin, Italy \quad
$^{2}$Eindhoven University of Technology, The Netherlands\\[0.2em]
$^{3}$LTCI, T\'el\'ecom Paris, Institut Polytechnique de Paris, France\\
{\tt\small riccardo.renzulli@unito.it}
}

\begin{document}
\maketitle
\begin{abstract}
Vision-Language Models (VLMs) have demonstrated impressive capabilities across different tasks, but their computational cost is dominated by the large number of visual tokens fed to the language model. Existing token reduction methods rely on attention-based scores or pairwise similarity, without an explicit semantic representation of each token. We introduce TORINO (TOken Reduction via Interpretable coNcept Overlap), a plug-and-play framework for adaptive visual token reduction in VLMs that requires no fine-tuning of the underlying model. TORINO leverages Sparse Autoencoders (SAEs) to project visual tokens into an interpretable latent space where token relationships can be analyzed through shared concept activations. Specifically, we define concept overlap as the degree of agreement between active SAE latents and use it to group tokens that share semantic content. Reduction within each group is then performed by either pruning or merging, providing a unified framework that preserves semantically important visual information while removing redundancy. Unlike fixed-budget approaches, TORINO dynamically adapts the
reduction rate to input complexity, allowing different images to retain different numbers of tokens. Experiments  across multiple vision-language benchmarks show that TORINO achieves favorable efficiency-accuracy
trade-offs, reducing the number of visual tokens with minimal performance loss. Source code released at \url{https://eidoslab.github.io/TORINO}.
\end{abstract}
    
\section{Introduction}
\label{sec:intro}
\begin{figure}[t]
\centering
\includegraphics[width=\linewidth]{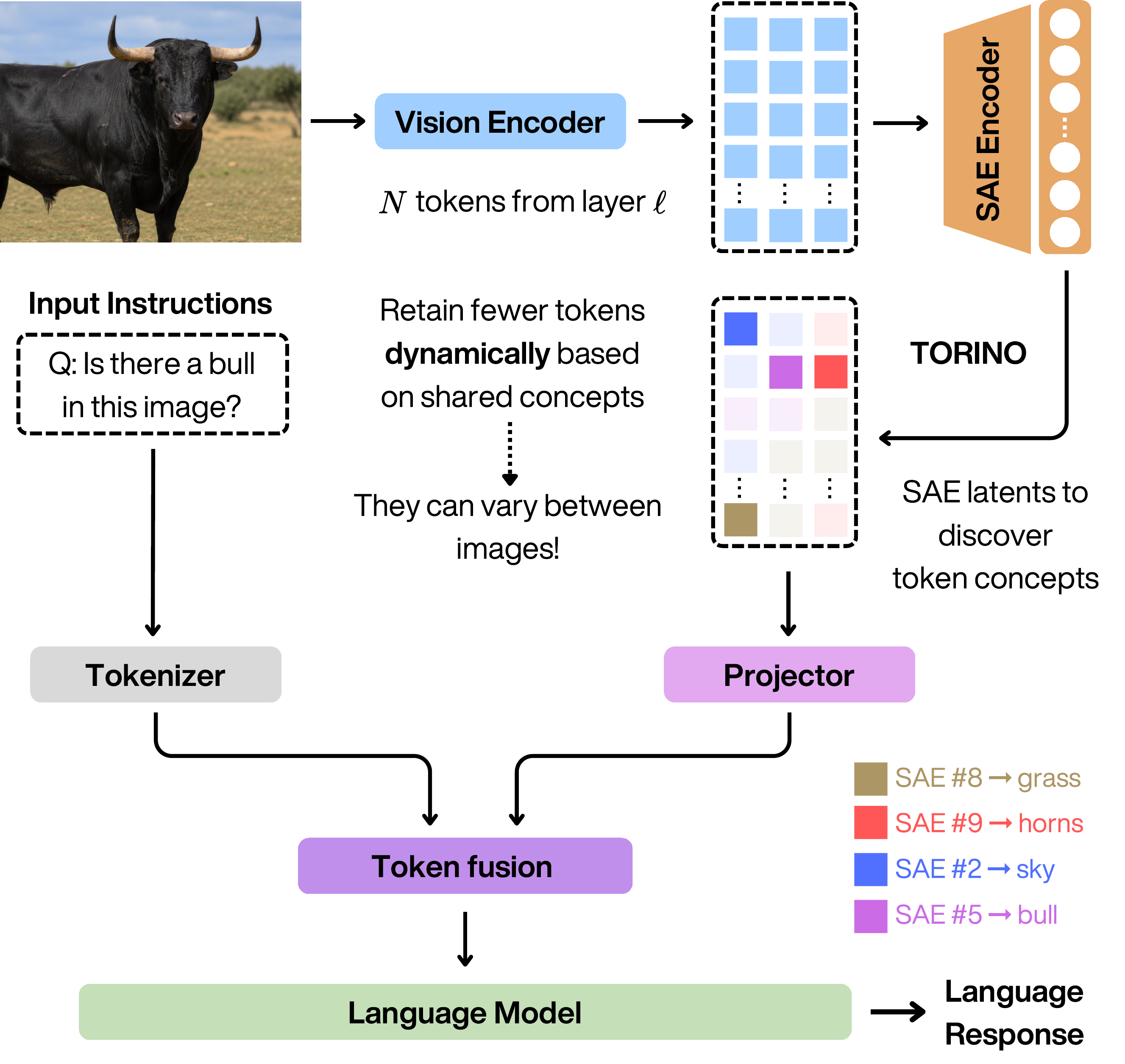}
\caption{TORINO reduces visual tokens dynamically by grouping patches according active
  SAE concept latents. The number of output tokens adapts automatically to image
  complexity: simple, uniform images collapse into fewer groups and are compressed
  more aggressively than richly structured ones.}
\label{fig:teaser}
\end{figure}

Vision-Language Models (VLMs)~\cite{liu2023llava, flamingo2022, Qwen-VL, dai2023instructblip},
have achieved remarkable performance across a wide range of multimodal tasks,
from visual question answering to image captioning and document understanding.
Their success, however, comes at a high computational cost.
In architectures such as LLaVA~\cite{liu2023llava}, a vision encoder
(typically a CLIP ViT~\cite{radford2021learning}) outputs hundreds of patch tokens per image,
all of which are passed to a large language model (LLM).
This token sequence constitutes the dominant bottleneck:
attention in the LLM scales quadratically with sequence length,
making visual token count a critical factor in both inference latency and memory footprint.
 
To address this, a growing body of work proposes to reduce the number of visual tokens
before they reach the LLM, via either \emph{pruning}, discarding tokens deemed
uninformative, or \emph{merging}, collapsing redundant tokens into
aggregates~\cite{bolya2023token, Shang_2025_ICCV, onemoretoken, fang2026prune, lu2025toma, arif2025hired, Alvar_2025_CVPR}.
Despite considerable progress, these methods share a fundamental limitation:
token importance is assessed through proxy signals such as attention scores,
CLS-token similarity, or raw embedding distance.
None of these signals corresponds to an explicit semantic representation of
\emph{what concept} each token encodes.
As a result, tokens that are visually dissimilar but semantically redundant may be incorrectly
retained, while tokens sharing abstract concepts may be kept separately.
 
Mechanistic interpretability offers a complementary perspective.
Sparse Autoencoders (SAEs)~\cite{bricken2023monosemanticity, bereska2024mechanistic}
have emerged as a principled tool for decomposing polysemantic
activations of neural networks into sparse, monosemantic latent features,
each tracking a distinct human-interpretable concept.
Recent work~\cite{pach2026sparse} has extended SAEs to vision-language encoders, revealing rich concept structure within
patch-level token representations.
Historically, such tools have been used purely for post-hoc analysis.
The community is now increasingly asking whether interpretability
tools can be \emph{repurposed} to guide functional decisions, what
has been termed pragmatic
interpretability~\cite{nanda2025pragmatic}.
 
We take this question to the visual token reduction setting.
We introduce TORINO (TOken Reduction via
Interpretable coNcept Overlap), a plug-and-play framework
that leverages pretrained Matryoshka BatchTopK
SAEs~\cite{bussmann2025learning}
to project visual tokens into an interpretable concept space and reduce them
based on shared SAE activations (see \cref{fig:teaser}).
Unlike attention or similarity-based methods, SAE latents can also capture
\emph{abstract} concepts
that transcend local appearance similarity.
TORINO groups tokens by their dominant active latent and then prunes or merges
within each group, exploiting the interpretable dictionary as the basis for
all reduction decisions.
Crucially, the grouping is \emph{content-adaptive}. Images with rich, diverse concepts form many small groups and undergo less reduction. Visually redundant images instead form fewer, larger groups and are compressed more aggressively, without requiring explicit complexity estimation.
For scenarios that require a predictable output length, we additionally
propose a fixed-budget variant that constrains TORINO to a target token count
while preserving the concept-guided grouping criterion.
 
\noindent Our main contributions are as follows. \fcirc{1} We introduce TORINO, the first SAE-based framework for visual token reduction in VLMs, requiring no fine-tuning of the base network and operating entirely at inference time via a fixed pretrained dictionary. \fcirc{2} We show that concept-based reduction is competitive with attention and similarity-based baselines across benchmarks in the moderate-reduction regime, where semantic redundancy is most exploitable, and provide an empirical analysis for this regime-specific advantage. 

\section{Preliminaries and Related Work}
\label{sec:relatedwork}

\noindent\textbf{SAEs background.}
Sparse Autoencoders (SAEs) have recently emerged as a powerful tool for mechanistic interpretability, enabling the decomposition of dense neural activations into sparse and often human-interpretable features. 
SAEs project activations into an \emph{overcomplete}
latent space in which each dimension is encouraged to be monosemantic.
Formally, let $\mathbf{v} \in \mathbb{R}^{d}$ be an embedding extracted
from a pretrained encoder.
An SAE is parameterized by an encoder matrix
$\mathbf{W}_{\!\mathrm{enc}} \in \mathbb{R}^{d \times \omega}$,
a decoder matrix
$\mathbf{W}_{\!\mathrm{dec}} \in \mathbb{R}^{\omega \times d}$,
and a shared bias $\mathbf{b} \in \mathbb{R}^{d}$,
where the latent width $\omega := \varepsilon d$ is controlled
by an expansion factor $\varepsilon \geq 1$.
The encoder and decoder maps are defined as
\begin{align}
  \varphi(\mathbf{v})
    &:= \sigma\!\left(\mathbf{W}_{\!\mathrm{enc}}^{\top}
        (\mathbf{v} - \mathbf{b})\right) \in \mathbb{R}^{\omega},
  \label{eq:sae_encoder}\\
  \psi(\mathbf{z})
    &:= \mathbf{W}_{\!\mathrm{dec}}^{\top}\mathbf{z} + \mathbf{b}
        \in \mathbb{R}^{d},
  \label{eq:sae_decoder}
\end{align}
where $\sigma{:}\,\mathbb{R}^{\omega}\!\to\mathbb{R}^{\omega}$
is a sparsity-inducing non-linearity, giving reconstruction
$\hat{\mathbf{v}} := \psi(\varphi(\mathbf{v}))$.
We write $\mathbf{z} := \varphi(\mathbf{v})$ for the sparse latent
vector and $z_{k} := [\mathbf{z}]_{k}$ for the activation of the
$k$-th latent feature.
SAEs are trained by minimising a reconstruction error, a sparsity
penalty, and an auxiliary loss that prevents feature collapse~\cite{gao2025scaling}:
\begin{equation}
  \mathcal{L}(\mathbf{v})
    := \mathcal{R}(\mathbf{v})
     + \lambda\,\mathcal{S}(\mathbf{v})
     + \mathcal{L}_{\mathrm{aux}}(\mathbf{v}),
  \label{eq:sae_loss}
\end{equation}
where $\mathcal{R}(\mathbf{v}) := \|\mathbf{v} - \hat{\mathbf{v}}\|_{2}^{2}$
is the mean squared reconstruction error.
The canonical ReLU
variant~\cite{bricken2023monosemanticity} sets
$\sigma\!:= \mathrm{ReLU}$ and
$\mathcal{S}(\mathbf{v}) := \|\mathbf{z}\|_{1}$.
The BatchTopK activation
function~\cite{bussmann2024batchtopk} improves upon standard
element-wise approaches by considering sparsity across batches
rather than individual examples, directly controlling the mean
number of active features $K := \|\mathbf{z}\|_{0}$
and removing the need to tune $\lambda$.
\emph{Matryoshka SAEs}~\cite{bussmann2025learning}
further impose a nested hierarchy on the feature dictionary.
This encourages feature subsets to capture the most
salient semantic concepts independently.
Matryoshka objective is decoupled from the activation
function and can be combined with any SAE variant, including BatchTopK~\cite{pach2026sparse}, which is the configuration we adopt in this work.
This hierarchical organization is particularly relevant for
visual token reduction: coarse latent concepts naturally group tokens
belonging to the same semantic region (\eg object-level),
while finer features capture subtler distinctions between spatially
adjacent tokens.
In TORINO, we leverage these properties to project visual tokens into
an interpretable concept space and quantify semantic overlap between
tokens via their SAE activation vectors~$\mathbf{z}$.
Token reduction decisions, whether pruning redundant tokens or merging
semantically similar ones, are thus guided by interpretable latent
concepts rather than raw embedding similarity.

\noindent\textbf{SAEs for VLMs.}
SAEs were originally developed to interpret the internal representations of LLMs~\cite{shu-etal-2025-survey}, and have since been applied to identify monosemantic features, analyze circuits, and steer model behavior.
Their application to the visual domain is more recent. Rao \etal~\cite{rao2024discover} use sparse autoencoders to automatically discover the visual concepts encoded by a model, naming them and training linear probes for task-agnostic classification.
Pach \etal~\cite{pach2026sparse} train Matryoshka BatchTopK SAEs on CLIP activations and show that the resulting features respond selectively to interpretable visual concepts, exhibiting high monosemanticity scores.
Shen \etal~\cite{shen2026vlsae} extend this to cross-modal representations, using a unified SAE concept set to interpret and improve vision-language alignment in VLMs.
Beyond interpretability, SAEs have been applied to concept erasure in diffusion models~\cite{cywinski2025saeuron,cassano2026saemnesia} and to reduce object hallucinations in VLMs by enhancing task-relevant visual features~\cite{Park_2026_WACV}.
TORINO is, to our knowledge, the first method to exploit the monosemantic structure of SAE features to guide token reduction at inference time.

\noindent\textbf{Token reduction methods.} 
Existing visual token reduction approaches broadly follow two strategies: pruning and merging. Early training-free merging methods such as ToMe \cite{bolya2023token} group tokens according to embedding similarity. GTP-ViT \cite{xu2024gtp} instead formulates reduction as graph-based token summarization, propagating information from less important tokens to spatially and semantically connected retained tokens. FOLDER \cite{Wang_Folder2025_ICCV} extends the merging paradigm through iterative bipartite matching and shows that, at mild-to-moderate reduction rates, merging generally preserves information better than directly dropping matched tokens. In VLMs, DyMU~\cite{wang2025dymu} also extends ToMe to combine dynamic token merging with virtual unmerging in a training-free and plug-and-play manner. LLaVA-PruMerge \cite{Shang_2025_ICCV} selects tokens through CLS-to-patch attention, clusters them using key similarity, and merges each cluster, while VisionZip \cite{Yang_2025_CVPR} identifies attention-dominant tokens and merges redundant tokens according to feature similarity. HiRED~\cite{arif2025hired} instead partitions the image hierarchically and allocates regional token budgets using CLS attention. More recent pruning methods support more aggressive compression: DivPrune \cite{Alvar_2025_CVPR} maximizes the minimum pairwise distance among retained tokens to preserve diversity, whereas PruneSID \cite{fang2026prune} jointly balances importance and diversity and optionally assigns image-dependent token budgets using an information score derived from global token similarity.
These methods nevertheless infer redundancy from attention, raw token similarity, or global statistics. In contrast, TORINO measures agreement in an SAE-derived concept space, allowing tokens with different local appearances but shared latent semantics to be identified as redundant. Moreover, its dynamic variant does not first estimate an image complexity score or assign a token budget: the retained length emerges directly from the number of concept groups activated by each image. The same semantic grouping supports both pruning and merging, enabling us to study their complementary behavior within a unified framework.
\section{Method}
\label{sec:method}

\begin{figure*}[t]
\centering
\includegraphics[width=1\linewidth]{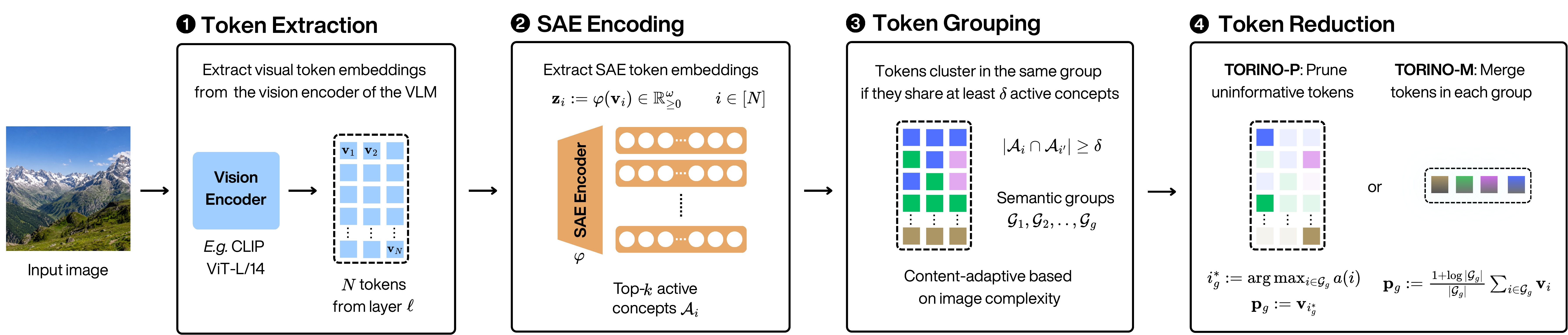}
\caption{TORINO pipeline. Visual patch tokens are extracted at layer $\ell$ of the frozen
  vision encoder, projected into an interpretable SAE concept space, grouped by shared
  active latents, and then pruned or merged within each group before being passed to the
  language model.}
\label{fig:pipeline}
\end{figure*}

TORINO operates entirely at inference time and requires no fine-tuning of
the underlying model.
Given a VLM whose vision encoder $f$ produces a sequence of
$N$ patch embeddings $\{\mathbf{v}_i\}_{i=1}^{N}$ at a chosen intermediate
layer $\ell$, TORINO proceeds in three stages:
(\emph{i}) encode each token via a pretrained SAE to obtain a sparse
concept vector,
(\emph{ii}) group tokens by shared active concepts,
(\emph{iii}) reduce each group to a single representative token by pruning
or merging.
An optional fixed-budget variant caps and pads the output to a
prescribed length.
The full pipeline is illustrated in \cref{fig:pipeline}.

\subsection{SAE Feature Extraction}
\label{sec:method:sae}

Let $f$ be a frozen VLM vision encoder with $L$ transformer blocks.
For a given image $\mathbf{x}$, we extract the intermediate activations at block $\ell < L$:
\begin{equation}
  \{\mathbf{v}_i\}_{i=1}^{N} = f^{(\ell)}(\mathbf{x}),
  \qquad \mathbf{v}_i \in \mathbb{R}^d.
  \label{eq:token_extract}
\end{equation}
An SAE $(\varphi, \psi)$ then can be trained on these activations to yield a
sparse decomposition into an overcomplete concept dictionary (as defined in
\cref{sec:relatedwork}).
Specifically, we adopt the Matryoshka BatchTopK SAE variant, which
simultaneously enforces batch-level sparsity and a nested concept hierarchy.
As shown by Pach~\etal~\cite{pach2026sparse}, Matryoshka SAEs achieve strictly higher
monosemanticity scores than vanilla or standard BatchTopK SAEs at equal
expansion factors, making their features the most suitable basis for
concept-guided token grouping.
At inference time, each token embedding is encoded as:
\begin{equation}
  \mathbf{z}_i := \varphi(\mathbf{v}_i) \in \mathbb{R}^{\omega}_{\geq 0},
  \qquad i \in [N],
  \label{eq:sae_encode}
\end{equation}
where $z_{ij} := [\mathbf{z}_i]_j$ denotes the activation of the $j$-th
concept feature for token $i$.
We additionally define the \emph{peak activation score} of a token as
\begin{equation}
  a(i) := \|\mathbf{z}_i\|_{\infty} = \max_{j \in [\omega]} z_{ij},
  \label{eq:peak_act}
\end{equation}
which serves as a unified concept importance measure.
In our experiments we use a publicly available pretrained
SAE~\cite{pach2026sparse}; details are given in \cref{sec:experiments}.

\subsection{Concept-Guided Token Grouping}
\label{sec:method:grouping}

Given sparse latent vectors $\{\mathbf{z}_i\}_{i=1}^N$, we group tokens that share dominant SAE concepts.
Two hyperparameters govern the grouping:
$k \in \mathbb{Z}_{>0}$, the number of top-active concepts retained
per token, and $\delta \in [1, k]$, the minimum concept overlap
required for two tokens to be co-grouped.
For each token $i$, let
\begin{equation}
  \mathcal{A}_i := \operatorname{argtop}_{k}(\mathbf{z}_i)
    \subset [\omega], \quad |\mathcal{A}_i| = k,
  \label{eq:act_set}
\end{equation}
be the index set of the $k$ largest entries of $\mathbf{z}_i$.
We construct an undirected \emph{token graph}
$\mathcal{H} = ([N], \mathcal{E})$ where an edge exists between tokens $i$
and $i'$ if and only if they share at least $\delta$ active concepts:
\begin{equation}
  (i, i') \in \mathcal{E}
    \iff |\mathcal{A}_i \cap \mathcal{A}_{i'}| \geq \delta.
  \label{eq:edge}
\end{equation}
The \emph{concept groups} $\{\mathcal{G}_g\}_{g=1}^{G_{\mathrm{dyn}}}$
are then defined as the connected components of $\mathcal{H}$, found via union-find.
Edge construction uses an inverted index over active concepts when $\delta=1$
($\mathcal{O}(kN)$), and a pairwise comparison over all token pairs otherwise
($\mathcal{O}(k^2N^2)$); both are negligible in practice for the patch counts
used in our experiments ($N=576$, $k\in\{1,2,3\}$).
Notably, the number of groups $G_{\mathrm{dyn}}$ is determined entirely by the image
content: semantically homogeneous images collapse into few large groups,
while richly structured images yield many small ones.
The pair $(k, \delta)$ jointly controls the granularity of grouping
by determining how readily tokens acquire edges in $\mathcal{H}$.
To make this concrete, consider two tokens $i$ and $i'$ whose dominant
concepts differ ($j^*(i) = c_1 \neq c_2 = j^*(i')$) but which
share a common secondary active concept $c_3$, so that
$\mathcal{A}_i = \{c_1, c_3\}$ and $\mathcal{A}_{i'} = \{c_2, c_3\}$
for $k = 2$:
\begin{itemize}[leftmargin=1.5em, itemsep=2pt, topsep=3pt]
  \item $(k{=}1,\;\delta{=}1)$:\enspace
        $(i,i') \notin \mathcal{E}$, since $c_1 \neq c_2$.
        Groups are disjoint argmax buckets; $G_{\mathrm{dyn}}$ equals
        the number of distinct dominant concepts active in the image.

  \item $(k{=}2,\;\delta{=}1)$:\enspace
        $(i,i') \in \mathcal{E}$, since
        $|\{c_1,c_3\} \cap \{c_2,c_3\}| = 1 \geq \delta$.
        Tokens sharing any top-$2$ concept are connected; via
        transitivity this merges argmax buckets that co-activate nearby
        dictionary atoms, yielding \emph{fewer and larger} groups than
        the $k{=}1$ baseline.

  \item $(k{=}2,\;\delta{=}2)$:\enspace
        $(i,i') \notin \mathcal{E}$, since
        $|\{c_1,c_3\} \cap \{c_2,c_3\}| = 1 < 2 = \delta$.
        Only tokens whose top-$2$ concept sets are identical share an
        edge; this is \emph{strictly stricter} than the $k{=}1$
        condition and yields \emph{more and smaller} groups.
\end{itemize}
In general, a higher ratio $\delta/k$ enforces tighter semantic
agreement, increasing $G_{\mathrm{dyn}}$ and so finer-grained
concept groups; a lower ratio permits looser affiliation, fusing more
tokens per group and yielding more aggressive compression.
Both $k$ and $\delta$ are image-agnostic hyperparameters; their
effect on $G_{\mathrm{dyn}}$ is thus  content-driven.


\subsection{Token Reduction}
\label{sec:method:reduction}

Each group $\mathcal{G}_g$ is reduced to a single \emph{primary token} $\mathbf{p}_g$.
We propose two complementary strategies.

\noindent\textbf{TORINO-P (Pruning).}
For each group, the token with the highest peak activation is selected as
the representative:
\begin{equation}
  i_g^* := \arg\max_{i \in \mathcal{G}_g} a(i), \qquad \mathbf{p}_g := \mathbf{v}_{i_g^*}.
  \label{eq:torino_p}
\end{equation}
All other tokens in the group are discarded.
TORINO-P retains original patch embeddings without modification,
which preserves the signal fidelity expected by the LLM's visual projection
layer.

\noindent\textbf{TORINO-M (Merging).}
Tokens within each group are aggregated into a single virtual token via a
log-size-rescaled mean:
\begin{equation}
  \mathbf{p}_g :=
    \frac{1 + \log|\mathcal{G}_g|}{|\mathcal{G}_g|}
    \sum_{i \in \mathcal{G}_g} \mathbf{v}_i,
  \label{eq:torino_m}
\end{equation}
i.e.\ the plain average of the group's patch embeddings, rescaled by
$1 + \log|\mathcal{G}_g|$. The sublinear factor $1 + \log|\mathcal{G}_g|$~\cite{bolya2023token} compensates for the fact that larger
groups are increasingly dominated by mutually redundant patches, so a
flat average would otherwise under-represent large, highly homogeneous
concepts relative to small ones.
Merged tokens are virtual embeddings in $\mathbb{R}^d$ and are fed to the
LLM's projection layer identically to real patch tokens.

\noindent In both cases, the $G_{\mathrm{dyn}}$ primary tokens
$\{\mathbf{p}_g\}_{g=1}^{G_{\mathrm{dyn}}}$ are passed to the language model,
replacing the original sequence of $N$ tokens.
The output length $G_{\mathrm{dyn}} \leq N$ is content-adaptive: no
explicit reduction ratio is specified, the model naturally reduces
more aggressively on visually redundant images.

\subsection{Fixed-Budget Variant}
\label{sec:method:budget}

The dynamic output length of TORINO can complicate batch processing and
fair comparison with fixed-ratio baselines.
We therefore introduce a \emph{fixed-budget} variant that constrains the
output to exactly $B$ tokens while preserving the concept-guided grouping
criterion.
The procedure augments the dynamic pipeline with two operations:
a \emph{truncation} step that trims excess groups before reduction, and a
\emph{padding} step that supplements the output when the image contains
fewer than $B$ distinct concept groups.

\noindent\textbf{Truncation.}
If the number of groups $G_{\mathrm{dyn}} > B$, we retain only the $B$
largest groups by patch count, which preferentially keeps concepts with
broad spatial support:
\begin{equation}
  \mathcal{K} :=
    \operatorname{argtop}_{B}\bigl(\{|\mathcal{G}_g|\}_{g=1}^{G_{\mathrm{dyn}}}\bigr),
  \label{eq:cap}
\end{equation}
and all tokens outside $\bigcup_{g \in \mathcal{K}} \mathcal{G}_g$ are
dropped.
After truncation, primary tokens $\{\mathbf{p}_g\}_{g \in \mathcal{K}}$ are then computed from the
retained groups using \cref{eq:torino_p} (TORINO-P) or
\cref{eq:torino_m} (TORINO-M).

\noindent\textbf{Padding.}
If $G_{\mathrm{dyn}} < B$, we supplement the $G_{\mathrm{dyn}}$ primary tokens with the
$B - G_{\mathrm{dyn}}$ highest-scoring tokens from the \emph{secondary pool}
$\mathcal{P}$, defined as:
\begin{equation}
  \mathcal{P} :=
  \begin{cases}
    \displaystyle\bigcup_{g \in \mathcal{K}} \mathcal{G}_g
      & \text{TORINO-M}, \\[6pt]
    \displaystyle\bigcup_{g \in \mathcal{K}} \mathcal{G}_g
      \setminus \{i_g^*\}_{g \in \mathcal{K}}
      & \text{TORINO-P}.
  \end{cases}
  \label{eq:pad_pool}
\end{equation}
For TORINO-M the primary tokens are virtual (merged) embeddings, so all
real patch tokens remain eligible for padding.
For TORINO-P the already-selected representative patches are excluded.
In both cases, pool members are ranked by $a(i)$ and the top $B - G_{\mathrm{dyn}}$ are
appended to the output, yielding exactly $B$ tokens.
Complete procedures for both variants
are provided as pseudocode in \cref{alg:torino_dynamic,alg:torino_budget} in the supplementary material.

\begin{table*}[t]
  \centering
  \small
  \setlength{\tabcolsep}{2.8pt}
  \caption{\textbf{Performance comparison on LLaVA-1.5-7B across 9 image understanding benchmarks.}
  SAE: Matryoshka BatchTopK 20, $\varepsilon=64$. Best performance per tier is in \textbf{bold}, second-best is \underline{underlined}. Relative is the macro-average of per-benchmark ratios against our reproduced baseline. Latency of the token reduction modules in isolation is reported in milliseconds.}
 \label{tab:results-llava-v1-5-7b-x64-cls-only}
   \begin{tabular*}{\textwidth}{
    @{\extracolsep{\fill}}
    lccccccccc|cc
    @{}
}
  \toprule
    Method & GQA & MMB$^{\text{EN}}$ & MMB$^{\text{CN}}$ & MME & POPE & SQA$^{\text{I}}$ & VQA$^{\text{T}}$ & VizWiz & MMVet & Rel.\ $\uparrow$ & Lat. (ms) $\downarrow$ \\
    \midrule
    \rowcolor{gray!15}
    \multicolumn{12}{c}{\textit{Upper Bound, 576 Tokens (100\%)}} \\
    \midrule
    Baseline & 61.9 & 64.0 & 57.9 & 1834.8 & 86.2 & 67.9 & 45.6 & 54.4 & 31.5 & 100.0\% & N/A \\
    \midrule
    \rowcolor{gray!15}
    \multicolumn{12}{c}{\textit{Retain $\sim$217 Tokens in Average ($\downarrow$ 62\%)}} \\
    \midrule
    Random & 59.1 & 61.2 & 54.1 & \underline{1772.6} & 84.7 & 67.7 & 34.3 & \underline{55.4} & 29.8 & 94.5\% & 13.6 $\pm$ 0.7 \\
    FOLDER & 59.4 & 60.7 & 52.7 & 1732.3 & \textbf{86.4} & \textbf{68.4} & 39.6 & 54.2 & 28.9 & 95.0\% & 16.6 $\pm$ 0.7 \\
    PruneSID & 59.1 & 62.1 & 55.8 & 1713.8 & 85.8 & 67.3 & 40.8 & 54.9 & 31.3 & 96.7\% & 64.6 $\pm$ 13.6 \\
    PruMerge & 58.3 & 62.3 & 56.4 & 1713.6 & 84.7 & 67.4 & 40.8 & \textbf{55.8} & 30.9 & 96.7\% & 71.2 $\pm$ 2.3 \\
    VisionZip & 59.4 & \underline{62.8} & 56.6 & 1747.6 & \underline{86.0} & \underline{67.8} & \textbf{44.4} & 53.8 & 32.4 & \underline{98.4\%} & 15.8 $\pm$ 0.4 \\
    TORINO-P & \underline{60.1} & 62.6 & 56.0 & 1749.5 & 85.7 & 67.2 & \underline{43.7} & 54.8 & 32.4 & 98.3\% & 27.6 $\pm$ 2.6 \\
    $\textrm{TORINO}_{FB}$-P & 59.8 & 62.5 & \textbf{56.8} & 1756.5 & 85.6 & 67.5 & 42.7 & 54.2 & 32.0 & 97.9\% & 28.1 $\pm$ 2.5 \\
    TORINO-M & \textbf{60.3} & 62.2 & 56.4 & \textbf{1778.8} & 85.3 & 67.6 & 43.6 & 54.6 & \textbf{33.1} & \textbf{98.7\%} & 26.4 $\pm$ 2.6 \\
    $\textrm{TORINO}_{FB}$-M & 59.8 & \textbf{63.0} & \underline{56.7} & 1740.4 & 85.1 & 67.4 & 42.5 & 53.9 & \underline{32.7} & 98.0\% & 26.6 $\pm$ 2.5 \\
    \midrule
    \rowcolor{gray!15}
    \multicolumn{12}{c}{\textit{Retain $\sim$129 Tokens in Average ($\downarrow$ 78\%)}} \\
    \midrule
    Random & 56.9 & 58.4 & 49.8 & 1673.8 & 82.2 & 67.3 & 28.0 & \textbf{57.0} & 26.9 & 89.6\% & 13.5 $\pm$ 0.5 \\
    FOLDER & 56.4 & 56.4 & 46.8 & 1641.0 & 82.5 & 67.1 & 32.1 & 56.0 & 29.3 & 90.0\% & 17.2 $\pm$ 0.7 \\
    PruneSID & 57.9 & 60.7 & 52.6 & 1724.7 & \textbf{84.8} & 67.5 & 40.1 & 56.1 & 27.4 & 94.3\% & 60.1 $\pm$ 18.4 \\
    PruMerge & 56.5 & \underline{61.9} & \underline{55.3} & 1677.5 & 74.7 & 67.9 & 39.3 & \underline{56.4} & \textbf{32.2} & 94.9\% & 48.2 $\pm$ 2.0 \\
    VisionZip & 57.8 & \textbf{62.9} & \textbf{56.2} & 1708.8 & 83.8 & \underline{68.3} & \textbf{42.0} & 55.5 & 31.2 & \underline{96.9\%} & 15.8 $\pm$ 0.5 \\
    TORINO-P & \textbf{58.8} & 61.0 & 55.2 & \textbf{1763.7} & \underline{84.0} & 68.0 & \underline{41.6} & 56.1 & \underline{31.6} & \textbf{97.1\%} & 29.6 $\pm$ 2.0 \\
    $\textrm{TORINO}_{FB}$-P & 58.5 & 61.9 & 55.2 & \underline{1745.9} & 83.8 & 68.0 & 40.3 & 55.8 & 30.9 & 96.4\% & 30.0 $\pm$ 1.8 \\
    TORINO-M & \underline{58.5} & 60.8 & 54.8 & 1742.6 & 83.6 & 67.8 & 41.4 & 55.5 & 30.0 & 96.0\% & 28.5 $\pm$ 2-0 \\
    $\textrm{TORINO}_{FB}$-M & 58.2 & 61.9 & 54.9 & 1715.7 & 83.2 & \textbf{68.4} & 40.0 & 55.6 & 28.7 & 95.2\% & 29.1 $\pm$ 1.7 \\
    \midrule
    \rowcolor{gray!15}
    \multicolumn{12}{c}{\textit{Retain $\sim$47 Tokens in Average ($\downarrow$ 92\%)}} \\
    \midrule
    Random & 53.9 & 51.5 & 39.9 & 1575.5 & 74.1 & 67.3 & 20.4 & 56.2 & 24.7 & 81.5\% & 14.0 $\pm$ 1.6 \\
    FOLDER & 51.0 & 46.2 & 32.6 & 1283.5 & 59.5 & 65.4 & 22.9 & 56.0 & 24.4 & 75.2\% & 21.0 $\pm$ 2.3 \\
    PruneSID & \textbf{56.4} & 59.6 & 50.6 & \textbf{1735.0} & \textbf{84.4} & 67.2 & 37.6 & \textbf{57.0} & 26.7 & \textbf{92.8\%} & 23.0 $\pm$ 1.8 \\
    PruMerge & 53.0 & \underline{59.9} & 51.1 & 1631.0 & 70.5 & 67.2 & \underline{37.9} & 56.7 & 29.4 & 90.9\% & 25.7 $\pm$ 0.8 \\
    VisionZip & 54.0 & \textbf{60.2} & \textbf{52.1} & 1600.1 & 74.6 & 67.1 & \textbf{40.5} & 55.6 & 27.3 & \underline{91.3\%} & 15.6 $\pm$ 0.2 \\
    TORINO-P & \underline{55.2} & 59.5 & \underline{51.4} & 1612.8 & \underline{77.4} & \underline{67.7} & 33.1 & 56.3 & \textbf{30.1} & 91.1\% & 21.8 $\pm$ 0.5 \\
    $\textrm{TORINO}_{FB}$-P & 54.7 & 59.2 & 50.6 & \underline{1648.4} & 77.2 & \textbf{67.9} & 32.5 & 56.2 & 28.8 & 90.4\% & 22.0 $\pm$ 0.3 \\
    TORINO-M & 54.9 & 58.0 & 50.2 & 1560.5 & 75.2 & 67.4 & 32.5 & \underline{56.7} & 29.4 & 89.6\% & 20.7 $\pm$ 0.6 \\
    $\textrm{TORINO}_{FB}$-M & 54.3 & 57.1 & 49.1 & 1625.7 & 75.0 & 67.2 & 31.8 & 56.2 & 28.0 & 88.7\% & 20.8 $\pm$ 0.4 \\
    \bottomrule
  \end{tabular*}
\end{table*}

\section{Experiments}
\label{sec:experiments}

\subsection{Experimental Setup}
\label{sec:experiments:setup}


\noindent\textbf{SAEs.}
We use the publicly available Matryoshka BatchTopK SAE~\cite{pach2026sparse}, trained on ImageNet-1K training-set CLS
activations extracted from CLIP ViT-L/14@336 at block~22 (post-MLP
residual).
The SAE uses $k=20$ active features per token and an expansion factor
$\varepsilon=64$, yielding a concept dictionary of $\omega=\varepsilon d=65{,}536$
features over a $d_{\mathrm{in}}=1{,}024$-dimensional input space.
All SAE weights are frozen; only the encoder $\varphi$ is executed at
inference time to produce the sparse latent $\mathbf{z}_i$ per patch. We adopt $\varepsilon=64$ because larger SAE expansion factors have been shown to produce more monosemantic features \cite{pach2026sparse}, which yield semantically coherent patch groups and are uniquely robust at aggressive compression; see \cref{sec:experiments:ablation} for a full sweep over $\varepsilon$.
We additionally report ablations over $\varepsilon\in\{1,2,4,8,16,64\}$
in \cref{sec:experiments:ablation} to assess sensitivity to dictionary size.

\noindent\textbf{VLMs.}
We evaluate on LLaVA-1.5-7B and LLaVA-1.5-13B~\cite{liu2023llava}, whose
vision encoder is a frozen CLIP ViT-L/14@336~\cite{radford2021learning}.
The encoder processes $336\times336$ resolution images, producing $N=576$ patch tokens
on a $24\times24$ spatial grid with embedding dimension $d=1{,}024$.
Token reduction is applied at the output of transformer block~22
(post-MLP residual stream), the layer whose activations are consumed by
LLaVA's visual projection layer.

\noindent\textbf{TORINO hyperparameters.}
We evaluate TORINO at three configurations:
$(k,\delta)\in\{(1,1),\,(2,2),\,(3,3)\}$.
In the \emph{dynamic} setting no budget is imposed, so the output length
$G_{\mathrm{dyn}}$ is fully content-adaptive; the three $(k,\delta)$ values
naturally induce different average compression levels.
In the \emph{fixed-budget} setting the same three configurations are paired
with target budgets $B\in\{192,128,64\}$.

\noindent\textbf{Baselines.}
We compare against five token-reduction methods, all applied at the same
block~22 and requiring no fine-tuning:
Random (uniform random patch sampling),
FOLDER~\cite{Wang_Folder2025_ICCV},
PruneSID~\cite{fang2026prune},
PruMerge~\cite{Shang_2025_ICCV}, and
VisionZip~\cite{Yang_2025_CVPR}.

\noindent\textbf{Benchmarks and evaluation protocol.}
All methods are evaluated with VLMEvalKit~\cite{duan2024vlmevalkit} on
nine benchmarks spanning diverse multimodal understanding tasks:
GQA~\cite{hudson2019gqa},
MMBench (English and Chinese)~\cite{liu2024mmbench},
MME~\cite{fu2023mme},
POPE~\cite{pope},
ScienceQA (image split)~\cite{lu2022sqa},
TextVQA~\cite{singh2019textvqa},
VizWiz~\cite{gurari2018vizwiz}, and
MM-Vet~\cite{yu2024mm}.
Because these benchmarks report on incompatible scales, we aggregate with
a \emph{relative} score: the macro-average of per-benchmark ratios against
a reproduced full-token baseline (576 tokens, no reduction).

\subsection{Quantitative Results}
\label{sec:experiments:main}
\cref{tab:results-llava-v1-5-7b-x64-cls-only} reports results on LLaVA-1.5-7B using our primary SAE configuration, Matryoshka BatchTopK with expansion factor $\varepsilon=64$. 
In the dynamic setting, TORINO selects an image-dependent number of output tokens, allowing the reduction rate to adapt to image content. 
To isolate the contribution of this adaptive budget allocation, we also report TORINO$_{\mathrm{FB}}$ variants, which use a fixed number of tokens matched to the average budget of their corresponding dynamic counterpart within each reduction tier. 
Full comparisons at conventional fixed budgets of 192, 128, and 64 tokens are provided in \cref{sec:appendix:results} of the supplementary material. 
At moderate levels of reduction, both TORINO variants consistently offer the strongest overall trade-off between performance and efficiency. In particular, TORINO-M achieves the best relative score (with respect to vanilla LLaVA-1.5-7B) when retaining approximately 217 tokens (98.7\%), while TORINO-P performs best at approximately 129 tokens (97.1\%). 
Comparing dynamic and fixed-budget variants at matched average budgets further indicates that adapting the retained-token count to the input can improve performance, especially in the moderate reduction regime. In the most aggressive setting, PruneSID attains the highest relative score, whereas TORINO-P still achieves similar results to VisionZip.
The last column in \cref{tab:results-llava-v1-5-7b-x64-cls-only} reports the latency of the token reduction module in isolation, measured on an NVIDIA A40 GPU across 50 randomly sampled POPE images after 10 warm-up forward passes (excluding the vision encoder and language-model forward passes). Although Random, Folder and VisionZip have the lowest overhead, TORINO-P and TORINO-M remain substantially faster than PruneSID and PruMerge at the two moderate reduction levels. For example, at approximately 217 retained tokens, TORINO-M requires 26.43\,ms, while PruneSID 64.56\,ms and PruMerge 71.18\,ms.

\noindent\textbf{Pruning vs.\ merging.}
TORINO-P and TORINO-M achieve comparable relative scores at low-to-moderate
compression, with TORINO-P pulling ahead as compression increases.
This pattern is consistent with the log-size rescaling in TORINO-M
(\cref{sec:method:reduction}): the sublinear factor partially compensates for
information loss in large groups, but at very high compression individual
groups grow large enough that even rescaled averages blur fine-grained detail. 

Complete results for full efficiency measurements, other expansion factors, additional random seeds, LLaVA-1.5-13B and LLaVA-NeXT  are deferred to \cref{sec:appendix:results} of the supplementary material.

\subsection{Per-image Compression Dynamics}
\label{sec:experiments:dynamics}
\begin{figure}[h]
  \centering
  \includegraphics[width=1\columnwidth]{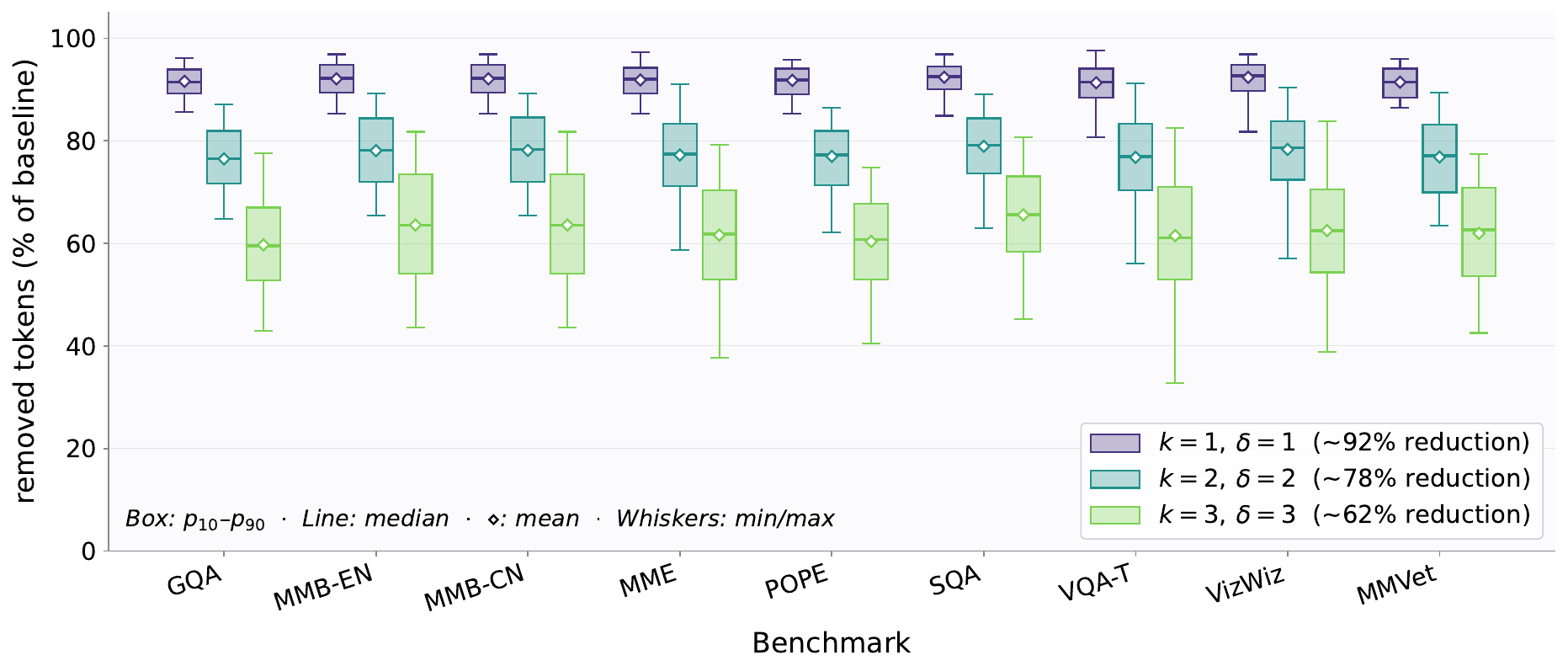}
  \caption{\textbf{Per-image distribution of removed tokens} for TORINO in the
    dynamic setting, on the Matryoshka BatchTopK $\varepsilon=64$ SAE, across
    the nine LLaVA-1.5-7B benchmarks. Each box summarises the distribution of
    per-image removal percentages within a benchmark: edges mark the 10th and
    90th percentiles (covering 80\% of images), the central line is the median,
    the diamond is the mean, and whiskers are the minimum and maximum.}
  \label{fig:torino-dynamic-distribution}
\end{figure}

\cref{tab:results-llava-v1-5-7b-x64-cls-only} exposes \emph{average} compression, but the dynamic setting
is best understood image-by-image across different benchmarks.
\cref{fig:torino-dynamic-distribution} reports the per-image distribution
of removed tokens for each $(k,\delta)$ configuration across the nine LLaVA-1.5-7B
benchmarks. The three $(k,\delta)$ configurations induce distinct compression regimes, but each one also displays substantial \emph{within-benchmark} spread, namely the visual signature of content-adaptivity. The distribution is identical for TORINO-P and TORINO-M, which differ only in the per-group reduction step and emit the same $G_{\mathrm{dyn}}$ tokens per image. Two observations emerge.

\noindent\textbf{Content-adaptivity is better at moderate sparsity ratios.}
The 10th and 90th percentiles
(covering 80\% of images) spread grows with $(k,\delta)$: at $(k=1,\delta=1)$
the boxes are narrow ($\sim$5--8 pp), while at $(k=3,\delta=3)$ the spread
reaches $\sim$20--25 pp: the same benchmark can retain anywhere from $\sim$115
to $\sim$330 tokens depending on image content.
This emergent budget explains the accuracy pattern in
\cref{tab:results-llava-v1-5-7b-x64-cls-only}: at $\sim$78\% reduction
the per-image budget is wide enough to cover the primary concept groups, giving
TORINO a $+1.6$--$+2.8$ pp lead over PruneSID and PruMerge. At $\sim$92\%
reduction the lower whiskers fall below 14 retained tokens for some images,
making the SAE partition too coarse.

\noindent\textbf{Cross-benchmark drift is consistent across regimes.}
The benchmarks order themselves in a stable way: SQA receives the largest
average compression at every $(k,\delta)$ value (tied with VizWiz at
$(k=1,\delta=1)$, both 13.1$\times$), while GQA, POPE, and VQA-T consistently
sit at the low end of the reduction spectrum.
This ordering tracks image complexity-benchmarks with simpler or more uniform
scenes (e.g.\ classroom diagrams in SQA) yield fewer distinct concept groups,
whereas images with denser visual content (GQA's Visual Genome scenes, POPE's
open-domain photographs, and VQA-T's text-in-scene images) activate more groups
and retain proportionally more tokens.
Crucially, the same benchmark-specific budget is recovered by TORINO without
any per-benchmark tuning, the SAE features alone provide the signal.

\subsection{Qualitative Results}
\label{sec:experiments:qualitative}

\begin{figure}[h]
  \centering
  \includegraphics[width=\columnwidth]{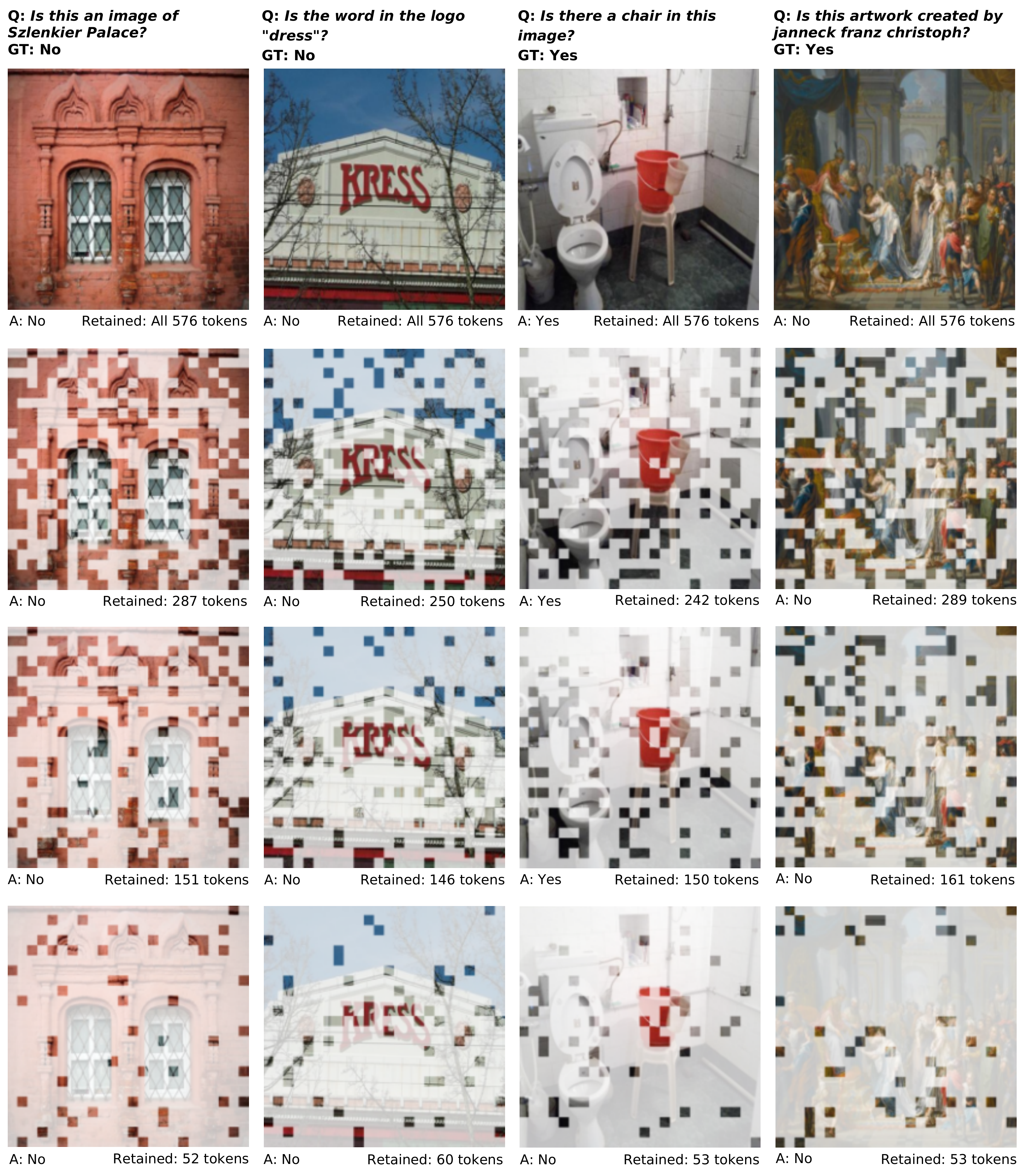}
  \caption{\textbf{Qualitative results for TORINO-P} ($\varepsilon=64$) on four
    MME examples. Rows correspond to base model and grouping configurations
    $(k,\delta)\in\{(3,3),(2,2),(1,1)\}$ (top to bottom); retained tokens are
    shown in full colour, pruned tokens are faded. Columns~1--2 show successful
    compression; column~3 illustrates a small-object failure at
    extreme sparsity; column~4 shows a baseline failure inherited from the base
    VLM, independent of token count.}
  \label{fig:qualitative}
\end{figure}

\cref{fig:qualitative} shows visualization on four MME
examples the base model LLaVA-1.5-7B without any reduction method and TORINO-P ($\varepsilon=64$) across three $(k,\delta)$ configurations.
In columns~1 and~2, the model answers correctly at all compression tiers: the
retained patches concentrate on the building façade and the signage characters,
which form large, spatially coherent concept groups that the SAE consistently
identifies as salient and preserves even at $\sim$52--60 tokens.
Column~3 illustrates a characteristic failure at extreme compression: the chair
is correctly identified at $\sim$242 and $\sim$150 tokens, but at $\sim$53
tokens its small concept group is pruned away and the model answers incorrectly.
This failure mode is inherent to any content-adaptive method at very high
reduction: small, spatially isolated objects form compact groups that are
outweighed by larger regions and fall below the retention threshold.
Column~4 shows a different failure: the artwork attribution question
(\emph{Is this artwork created by janneck, franz christoph?}) is answered
incorrectly at all three compression levels, including at $\sim$289 retained
tokens.
This is a fine-grained factual question that LLaVA-1.5-7B cannot answer
regardless of token count; TORINO does not introduce the error, which is
inherited from the base model's knowledge limitations.

\subsection{Ablation Studies}
\label{sec:experiments:ablation}

\noindent\textbf{Are SAEs Necessary?}
A natural question is whether the SAE is itself responsible for TORINO's gains, or
whether the grouping rule alone, applied directly to the raw patch embeddings
$\mathbf{v}_i$ at the same layer, is sufficient.
We isolate this by replacing the sparse latent $\mathbf{z}_i$ with the raw
embedding $\mathbf{v}_i$ and running the identical pipeline: top-$k$ feature
selection, edge formation via $\delta$-overlap, and connected-component grouping.
At matched compression ($\sim$47 retained tokens per image on average,
$\downarrow$92\% reduction), removing the SAE drops Relative score by 5.2
percentage points for the merging variant, from $89.60\%$ to $84.40\%$
(\cref{tab:ablation-nosae}).

\noindent\textbf{CLS and spatial tokens}.
Although the SAE was trained on CLS-token activations at block 22 of CLIP ViT-L/14,
by this late layer patch and CLS representations have undergone extensive
bidirectional attention mixing across all preceding blocks, substantially
narrowing the gap between their activation statistics.
To directly quantify the impact of training token choice, we also evaluate an SAE
trained on \emph{two randomly sampled spatial patch tokens per image}.
\Cref{tab:results-llava-v1-5-7b-fb-x64-cls-only} in the supplementary provides a controlled comparison
at matched fixed budgets ($B\in\{192,128,64\}$): the spatial SAE consistently
underperforms the CLS SAE, with the gap widening at higher compression
(TORINO-M: $-2.7$pp at $B{=}192$, $-7.7$pp at $B{=}128$, $-19.6$pp at $B{=}64$).
Dynamic-setting results for the spatial SAE are reported in
\cref{tab:results-llava-v1-5-7b-x64-random-k-2} (supplementary).
To mitigate this gap, training the spatial SAE on all patch tokens of a large vision dataset would expose the dictionary to substantially more visual diversity; we discuss this direction in \cref{sec:conclusion}.

\begin{table}[t]
  \centering
  \caption{\textbf{Impact of the SAE.} TORINO's grouping rule applied to raw
    patch embeddings (No SAE) versus SAE-encoded features ($\varepsilon=64$),
    at matched compression ($\sim$47 retained tokens, $\downarrow$92\%).
    Relative is the macro-average of per-benchmark ratios against the
    LLaVA-1.5-7B baseline.}
  \label{tab:ablation-nosae}
   \begin{tabular*}{.8\columnwidth}{
    @{\extracolsep{\fill}}
    lc
    @{}
}
    \toprule
    Method & Relative \\
    \midrule
    No SAE & 84.40\% \\
    TORINO-M ($\varepsilon=64$) & \textbf{89.60\%} \\
    \bottomrule
  \end{tabular*}
\end{table}

\begin{figure}[t]
  \centering
  \includegraphics[width=\linewidth]{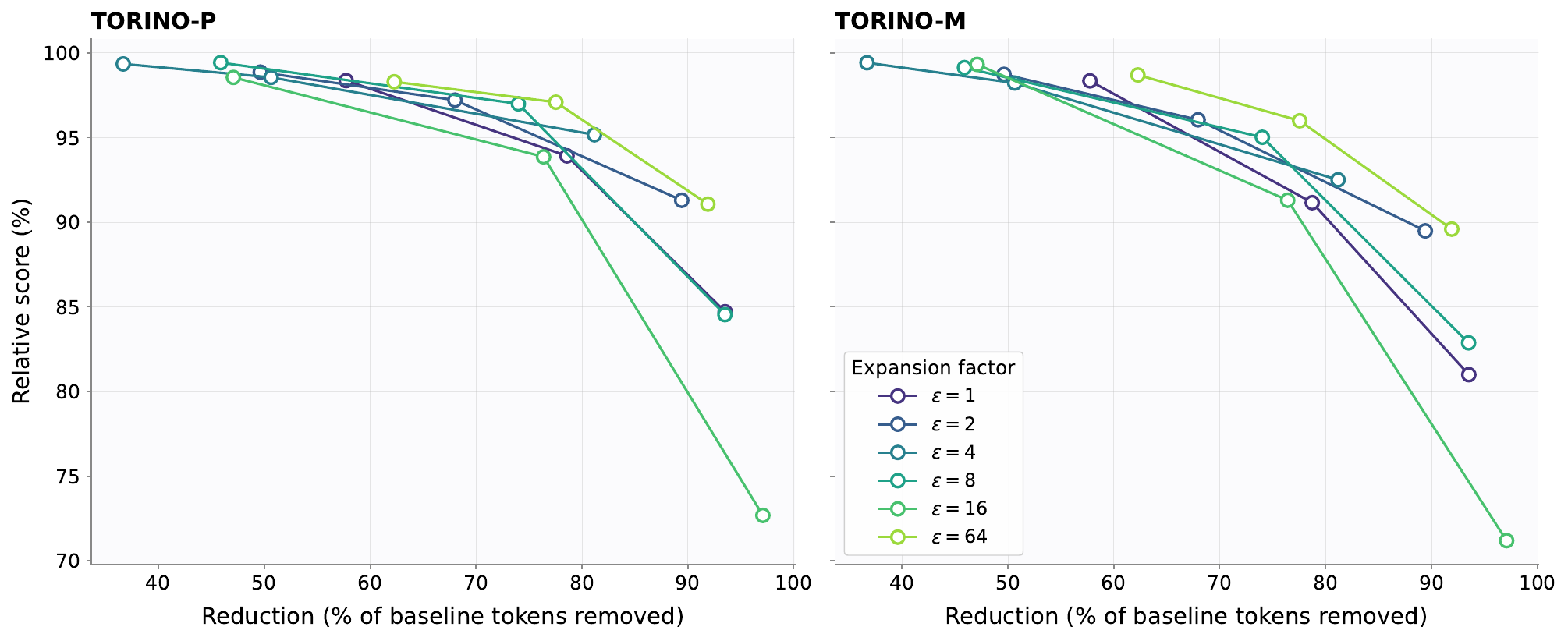}
  \caption{\textbf{Pareto trade-off across SAE expansion factors $\varepsilon$.}
    Each marker is one TORINO run; each curve connects  three grouping
    configurations $(k,\delta)\in\{(1,1),(2,2),(3,3)\}$.
    }
  \label{fig:ablation-epsilon}
\end{figure}

\noindent\textbf{SAE Expansion Factor.}
\cref{fig:ablation-epsilon} sweeps $\varepsilon\in\{1,2,4,8,16,64\}$
and plots each (variant, $\varepsilon$, $(k,\delta)$) triple on the
retention-accuracy Pareto plane. At moderate compression ($\downarrow$50--78\%), $\varepsilon=4$ traces the
dominant frontier, peaking at $99.35\%$ relative score for TORINO-P at
$(k=3,\delta=3)$, the highest single point across all configurations.
Smaller dictionaries ($\varepsilon\in\{1,2\}$) group semantically unrelated
patches due to polysemantic features; larger ones ($\varepsilon\geq 8$) fragment
well-connected concept regions into many small components, increasing variance
without improving fidelity.
At aggressive compression ($\downarrow$92\%, $\sim$47 retained tokens),
$\varepsilon=64$ becomes uniquely robust: TORINO-P reaches $91.07\%$ while the
next-best ($\varepsilon=8$) drops to $84.55\%$.
The exception is $\varepsilon=16$, which collapses to $72.7\%$ for TORINO-P at
$(k=1,\delta=1)$, the sharpest fall across all settings, likely caused by a
small number of high-frequency features dominating the argmax and collapsing
nearly all tokens into a single group.
The robustness of $\varepsilon=64$ is consistent with the monosemanticity
argument: highly specialised features produce groups whose members genuinely
share a visual concept, so even a handful of surviving groups carries
non-redundant image information.
Finally, TORINO-M leads TORINO-P at low-to-moderate reduction while TORINO-P
takes over at high reduction across the full $\varepsilon$ sweep, confirming
this pattern is not an artefact of any particular dictionary size.

\section{Conclusion}
\label{sec:conclusion}

We introduced TORINO, a plug-and-play framework for visual token reduction in
VLMs that exploits sparse autoencoders monosemantic latents.
By grouping tokens according to shared SAE activations and reducing within each
group via pruning or merging, TORINO achieves content-adaptive compression
without modifying any model weights.
Experiments on LLaVA models across nine benchmarks show that TORINO
performs well at moderate compression and remains competitive at extreme compression, trailing PruneSID and VisionZip by narrow margins.
The advantage at moderate compression also holds on LLaVA-NeXT's
higher-resolution, multi-crop setting.
However, applying TORINO to VLMs with vision encoders would require training a different SAE, a feasible step given the growing availability of open SAE tooling. Two directions offer promising paths forward.
First, the SAE used in this work was trained only on CLS-token
activations; training on the full set of spatial patch
tokens from a large vision dataset would give the encoder
direct exposure to the patch-level concept structure that TORINO relies on for
grouping, potentially improving both coverage and granularity of the resulting
feature dictionary.
Second, extending TORINO with an SAE trained jointly on visual and textual modalities, would enable query-conditioned grouping, where concept overlap is measured in a shared vision-language space and tokens
irrelevant to the query are pruned more aggressively, bringing interpretable
token reduction closer to the selective attention that humans apply when
answering image questions.

\section*{Acknowledgement}
The `Mechanistically-Grounded Adaptive AI' action has received funding from the European Union, via the oc4-2025-TES-02 issued and implemented by the ENFIELD project, under the grant agreement No 101120657. The `Pruning-Aware Adapters for Bitrate and Complexity Scalable LIC Models' action has received funding from the European Union, via the oc3-2025-TES-01 issued and implemented by the ENFIELD project, under the grant agreement No 101120657. We acknowledge the CINECA award under the ISCRA initiative, for the availability of high performance computing resources and support.
{
    \small
    \bibliographystyle{ieeenat_fullname}
    \bibliography{main}
}


\clearpage
\appendix

\section{Pseudocode for TORINO}
\label{app:algorithms}

We provide pseudocode for both the dynamic (\cref{alg:torino_dynamic})
and fixed-budget (\cref{alg:torino_budget}) variants of TORINO.
The dynamic variant produces a variable number of output tokens
$G_{\mathrm{dyn}}$ determined purely by image content, while the
fixed-budget variant enforces exactly $B$ output tokens via the
truncation and padding operations described in \cref{sec:method:budget}.
Both algorithms share the same SAE encoding and grouping front-end
(lines 1--8 of each);
the fixed-budget variant adds group selection (truncation, lines 9--12)
and pool-based padding (padding, lines 18--23).

\begin{algorithm}[H]
\caption{TORINO --- Dynamic Token Reduction}
\label{alg:torino_dynamic}
\begin{algorithmic}[1]
\Require Token embeddings $\{\mathbf{v}_i\}_{i=1}^{N}$;
         SAE encoder $\varphi$;
         grouping parameters $k \in \mathbb{Z}_{>0}$,
         $\delta \in [1, k]$;
         mode $\in \{\mathrm{P},\,\mathrm{M}\}$
\Ensure  Reduced sequence of $G_{\mathrm{dyn}}$ tokens
         (content-adaptive length)
\LineComment{--- Stage 1: SAE encoding}
\For{$i = 1$ \textbf{to} $N$}
  \State $\mathbf{z}_i \;\leftarrow\; \varphi(\mathbf{v}_i)$
  \State $a(i) \;\leftarrow\; \|\mathbf{z}_i\|_\infty$
         \hfill$\triangleright$\;\textit{peak activation score}
\EndFor
\LineComment{--- Stage 2: concept-guided grouping}
\For{$i = 1$ \textbf{to} $N$}
  \State $\mathcal{A}_i \;\leftarrow\;
         \operatorname{argtop}_k(\mathbf{z}_i)$
         \hfill$\triangleright$\;\textit{top-$k$ active concept indices}
\EndFor
\State Build token graph $\mathcal{H} = ([N],\mathcal{E})$
       where $(i,i')\in\mathcal{E}$\; iff
       \;$|\mathcal{A}_i\cap\mathcal{A}_{i'}|\ge\delta$
\State $\{\mathcal{G}_g\}_{g=1}^{G_{\mathrm{dyn}}} \;\leftarrow\;
       \textsc{ConnectedComponents}(\mathcal{H})$
       \hfill$\triangleright$\;\textit{union-find, $\mathcal{O}(N)$}
\LineComment{--- Stage 3: reduction}
\For{$g = 1$ \textbf{to} $G_{\mathrm{dyn}}$}
  \If{mode $= \mathrm{M}$}
    \State $\mathbf{p}_g \;\leftarrow\;
           \displaystyle\frac{1+\log|\mathcal{G}_g|}{|\mathcal{G}_g|}
           \sum_{i\in\mathcal{G}_g} \mathbf{v}_i$
           \hfill$\triangleright$\;\textit{log-size-rescaled merge}
  \Else\quad\{mode $= \mathrm{P}$\}
    \State $i_g^* \leftarrow \arg\max_{i\,\in\,\mathcal{G}_g}\,a(i)$;\enspace
           $\mathbf{p}_g \leftarrow \mathbf{v}_{i_g^*}$
           \hfill$\triangleright$\;\textit{highest-scoring representative}
  \EndIf
\EndFor
\State \Return $\{\mathbf{p}_g\}_{g=1}^{G_{\mathrm{dyn}}}$
\end{algorithmic}
\end{algorithm}

\begin{algorithm}[H]
\caption{TORINO --- Fixed-Budget Token Reduction}
\label{alg:torino_budget}
\begin{algorithmic}[1]
\Require Token embeddings $\{\mathbf{v}_i\}_{i=1}^{N}$;
         SAE encoder $\varphi$;
         target budget $B \in \mathbb{Z}_{>0}$;
         grouping parameters $k$, $\delta$;
         mode $\in \{\mathrm{P},\,\mathrm{M}\}$
\Ensure  Reduced sequence of exactly $B$ tokens
\LineComment{--- Stages 1--2: encoding and grouping (same as \cref{alg:torino_dynamic})}
\For{$i = 1$ \textbf{to} $N$}
  \State $\mathbf{z}_i \leftarrow \varphi(\mathbf{v}_i)$;\quad
         $a(i) \leftarrow \|\mathbf{z}_i\|_\infty$
\EndFor
\For{$i = 1$ \textbf{to} $N$}
  \State $\mathcal{A}_i \leftarrow \operatorname{argtop}_k(\mathbf{z}_i)$
\EndFor
\State Build $\mathcal{H}$ and compute
       $\{\mathcal{G}_g\}_{g=1}^{G}$ as in lines 6--8 of
       \cref{alg:torino_dynamic}
\LineComment{--- Truncation: retain the $B$ largest groups}
\If{$G > B$}
  \State $\mathcal{K} \;\leftarrow\;
         \operatorname{argtop}_{B}\!\bigl(\{|\mathcal{G}_g|\}_{g=1}^{G}\bigr)$
         \hfill$\triangleright$\;\textit{keep concepts with most spatial support}
  \State $G \;\leftarrow\; B$
\Else
  \State $\mathcal{K} \;\leftarrow\; [G]$
\EndIf
\LineComment{--- Stage 3: primary token per kept group}
\For{$g \in \mathcal{K}$}
  \If{mode $= \mathrm{M}$}
    \State $\mathbf{p}_g \;\leftarrow\;
           \displaystyle\frac{1+\log|\mathcal{G}_g|}{|\mathcal{G}_g|}
           \sum_{i\in\mathcal{G}_g} \mathbf{v}_i$
  \Else\quad\{mode $= \mathrm{P}$\}
    \State $\mathbf{p}_g \;\leftarrow\; \arg\max_{i\in\mathcal{G}_g}\,a(i)$
  \EndIf
\EndFor
\LineComment{--- Padding: supplement with highest-scoring pool tokens}
\If{$G < B$}
  \State $\mathcal{P} \;\leftarrow\;
         \bigcup_{g\in\mathcal{K}}\mathcal{G}_g$
  \If{mode $= \mathrm{P}$}
    \State $\mathcal{P} \;\leftarrow\;
           \mathcal{P} \setminus \{\mathbf{p}_g \mid g \in \mathcal{K}\}$
           \hfill$\triangleright$\;\textit{primaries already selected}
  \EndIf
  \State Append top-$(B{-}G)$ tokens from $\mathcal{P}$
         ranked by $a(\cdot)$
\EndIf
\State \Return $\{\mathbf{p}_g\}_{g\in\mathcal{K}} \;\cup\; \text{pad tokens}$
\end{algorithmic}
\end{algorithm}

\section{Additional Results}
\label{sec:appendix:results}
Table~\ref{tab:efficiency} reports the full efficiency profile of TORINO
against five token-reduction baselines at three matched budgets. 
Measured on an NVIDIA A40 GPU across 50 randomly sampled POPE images after 10 warm-up forward passes and a $128$-token generation cap (the LLaVA-1.5 evaluation budget we used in the main paper). The vision
tower runs \emph{once} per sample for every row. Reducing $576$ tokens to $49$
cuts LLM prefill FLOPs by $12.7\times$ and the KV cache by $11.8\times$,
which translates into a measured $1.94\times$ prefill speedup and a
$1.70\times$ reduction in end-to-end inference latency. 
Inference Time minus Prefill Time is approximately one decode step in every row, confirming that all methods emitted answers of the same length and that the latency
differences are prefill-driven rather than an artefact of answer verbosity.
The comparison with prior work is clearest in the Prefill column. TORINO is
faster than both PruneSID and PruMerge at every budget, and the margin widens
as the budget grows: $-7$\,ms at $49$ tokens, $-44$\,ms at $137$ and
$-53$\,ms at $230$. The Red.\ column explains why. TORINO's overhead is a
single dense SAE encoder matmul, independent of the budget at $0.077$\,T in
every tier, whereas PruneSID's PSCA and pairwise-similarity stage and
PruMerge's attention recomputation both scale with the number of retained
tokens. At the $230$-token budget this is severe enough that both methods are
slower than performing \emph{no reduction at all} ($161$\,ms and $157$\,ms
against the unreduced baseline's $140$\,ms): their selection overhead exceeds
the prefill they save. Notably, TORINO is faster in wall-clock despite
spending more analytic FLOPs on reduction than either method, one wide
dense matmul runs near peak throughput, while a data-dependent stream of small
kernels does not, so FLOPs and latency diverge in both directions.
TORINO does not dominate the three cheapest baselines. Random, FOLDER, and
VisionZip are $8$--$25$\,ms, $3$--$18$\,ms, and $8$--$26$\,ms faster
respectively, which is expected: Random is a permutation and a gather with
no multiply--accumulates, FOLDER's bipartite matching costs $0.0002$\,T,
some $340\times$ less than the SAE encode, and VisionZip's attention-based
selection costs at most $0.0000271$\,T, over $2800\times$ less. That advantage is purely computational and is paid for in accuracy
(Table~\ref{tab:results-llava-v1-5-7b-x64-cls-only} in the main paper).
Two further observations. Decode Time is flat at $29.8$--$31.6$\,ms/token
across every row including the unreduced baseline, confirming that token
reduction buys time-to-first-token rather than throughput: the KV cache
shrinks by up to $11.8\times$, but per-step decoding is dominated by weight
loading rather than attention. Peak GPU memory is essentially unchanged.
\begin{table*}
\centering
\caption{\textbf{End-to-end efficiency on LLaVA-1.5-7B.} 
\label{tab:efficiency}
SAE: Matryoshka BatchTopK 20 $\varepsilon = 64$, CLS-only. \emph{FLOPs} and \emph{Red.} are analytic, in teraFLOPs per image; the
timing, KV-cache and memory columns are measured. \emph{FLOPs} counts the LLM prefill only. \emph{Red.} is the reduction method's own cost, charged separately. Decode Time is
per token. }
\resizebox{\textwidth}{!}{%
\begin{tabular}{l|cccccccc}
\toprule
 Method & \# Token & \shortstack{FLOPs\\(T)} & \shortstack{Red.\\(T)} & \shortstack{Inference Time\\(ms)} & \shortstack{Prefill Time\\(ms)} & \shortstack{Decode Time\\(ms/token)} & \shortstack{KV Cache\\(MB)} & \shortstack{GPU Memory\\(GB)} \\
\midrule
 baseline & 576 & 7.6 & 0 & 173 & 140 & 31.6 & 288.0 & 14.1 \\
\midrule
Random & 230 & 3.0 & 0 & 124 & 93 & 30.6 & 115.0 & 13.8 \\
FOLDER & 230 & 3.0 & 0.000213 & 129 & 97 & 30.3 & 115.0 & 13.8 \\
PruneSID & 230 & 3.0 & 0.00108 & 193 & 161 & 30.4 & 115.0 & 13.9 \\
PruMerge & 230 & 3.0 & 0.000953 & 188 & 157 & 30.4 & 115.0 & 13.8 \\
VisionZip & 230 & 3.0 & 0.0000271 & 122 & 91 & 30.3 & 115.0 & 13.8 \\
TORINO-P & 230 & 3.0 & 0.0773 & 139 & 108 & 30.5 & 115.0 & 14.0 \\
TORINO-M & 230 & 3.0 & 0.0773 & 139 & 108 & 30.4 & 115.0 & 14.0 \\
\midrule
Random  & 137 & 1.8 & 0 & 101 & 70 & 30.1 & 68.5 & 13.8 \\
FOLDER & 137 & 1.8 & 0.000223 & 108 & 77 & 30.0 & 68.5 & 13.8 \\
PruneSID & 137 & 1.8 & 0.00092 & 171 & 139 & 30.2 & 68.5 & 13.9 \\
PruMerge & 137 & 1.8 & 0.000843 & 139 & 108 & 30.2 & 68.5 & 13.8 \\
VisionZip & 137 & 1.8 & 0.0000201 & 100 & 69 & 29.9 & 68.5 & 13.8 \\
TORINO-P & 137 & 1.8 & 0.0773 & 131 & 101 & 30.1 & 68.5 & 14.0 \\
TORINO-M & 137 & 1.8 & 0.0773 & 125 & 95 & 30.2 & 68.5 & 14.0 \\
\midrule
Random & 49 & 0.6 & 0 & 94 & 64 & 30.1 & 24.5 & 13.8 \\
FOLDER & 49 & 0.6 & 0.000226 & 100 & 69 & 29.9 & 24.5 & 13.8 \\
PruneSID & 49 & 0.6 & 0.000764 & 108 & 79 & 29.8 & 24.5 & 13.8 \\
PruMerge & 49 & 0.6 & 0.00074 & 108 & 78 & 30.1 & 24.5 & 13.8 \\
VisionZip & 49 & 0.6 & 0.00000917 & 93 & 64 & 29.7 & 24.5 & 13.8 \\
TORINO-P & 49 & 0.6 & 0.0773 & 102 & 72 & 30.0 & 24.5 & 14.0 \\
TORINO-M & 49 & 0.6 & 0.0773 & 103 & 73 & 30.0 & 24.5 & 14.0 \\
\bottomrule
\end{tabular}
}
\end{table*}

Table~\ref{tab:results-llava-v1-5-13b-x64-cls-only} reports LLaVA-1.5-13B results with $\varepsilon=64$.
TORINO leads at both moderate-compression tiers, though the absolute gap over the strongest baseline narrows relative to 7B (e.g.\ TORINO-P leads by $0.54$pp at $\downarrow$62\% versus $\sim\!2$pp on 7B); at extreme compression ($\downarrow$92\%), PruneSID takes the lead ($91.74\%$) while TORINO-P remains second-best ($90.55\%$), the same ordering as on 7B.
This compression of inter-method differences is consistent with a stronger LLM backbone being more robust to imperfect token selection in general, an effect that benefits all reduction methods uniformly rather than TORINO specifically.

Table~\ref{tab:results-llava-v1-5-7b-fb-x64-cls-only} shows LLaVA-1.5-7B fixed-budget results with $\varepsilon=64$, including r2 spatial-SAE variants for direct budget-matched comparison.

Tables~\ref{tab:results-llava-v1-5-7b-x1-cls-only}--\ref{tab:results-llava-v1-5-7b-x16-cls-only}
report dynamic results on LLaVA-1.5-7B for $\varepsilon\in\{1,2,4,8,16\}$.

Table~\ref{tab:results-llava-v1-5-7b-x64-random-k-2} reports dynamic results for the SAE trained on two random spatial tokens per image. 

Furthermore, we report in Tab.~\ref{tab:results-llava-v1-6-7b-x64-cls-only} results on LLaVA-NeXT as modern high resolution and multi-crop VLM, where we confirm the message of TORINO performing well at moderate levels of reduction and remaining competitive at the most aggressive setting, trailing PruneSID and VisionZip by narrow margins. 

Tab.~\ref{tab:results-llava-v1-5-7b-x64-cls-only-randomseeds} reports Random with 5 seeds at the $\downarrow$62\% tier. Its relative score is stable across seeds, so TORINO's better performance in the moderate regime is not sampling noise. Random being competitive at moderate compression is expected: by block 22 of a 24-block ViT, every patch has undergone $\sim$22 rounds of self-attention and absorbed substantial global context (the same late-layer mixing we invoke in Sec. 4.5 to justify SAE transfer from CLS to patches), so at moderate reduction almost any $\sim$217-patch subset already covers most locally-distinct content by volume. This also explains why the gap to the best method widens with more aggressive compression.

\begin{table*}[t]
  \centering
  \caption{Performance comparison on LLaVA-1.5-13B across 9 image understanding benchmarks. SAE: Matryoshka BatchTopK 20 $\varepsilon = 64$, CLS-only. Best per tier in \textbf{bold}, second-best \underline{underlined}. Relative is the macro-average of per-benchmark ratios against our reproduced baseline.}
  \label{tab:results-llava-v1-5-13b-x64-cls-only}
  \resizebox{\textwidth}{!}{%
  \begin{tabular}{lcccccccccc}
    \toprule
    Method & GQA & MMB$^{\text{EN}}$ & MMB$^{\text{CN}}$ & MME & POPE & SQA$^{\text{I}}$ & VQA$^{\text{T}}$ & VizWiz & MMVet & Relative \\
    \midrule
    \rowcolor{gray!15}
    \multicolumn{11}{c}{\textit{Upper Bound, 576 Tokens (100\%)}} \\
    \midrule
    Baseline & 63.50 & 68.21 & 62.63 & 1787.33 & 88.52 & 72.38 & 48.91 & 56.14 & 35.70 & 100.00\% \\
    \midrule
    \rowcolor{gray!15}
    \multicolumn{11}{c}{\textit{Retain $\sim$217 Tokens in Average ($\downarrow$ 62\%)}} \\
    \midrule
    Random & 58.80 & 62.54 & 58.08 & \textbf{1781.10} & 86.49 & 70.95 & 36.04 & 55.26 & 33.40 & 93.12\% \\
    FOLDER & 59.80 & 65.81 & 59.97 & 1678.18 & \textbf{88.38} & 71.59 & 42.79 & 55.46 & 34.60 & 95.80\% \\
    PruneSID & 59.15 & 65.98 & 60.22 & 1740.80 & \underline{88.06} & 71.49 & 43.30 & 55.36 & \underline{34.80} & 96.26\% \\
    PruMerge & 59.22 & 65.98 & \textbf{61.34} & 1729.59 & 87.04 & \textbf{72.24} & 43.07 & \textbf{55.63} & \textbf{35.80} & 96.69\% \\
    TORINO-P & \textbf{60.13} & \underline{66.15} & \underline{60.40} & \underline{1753.61} & 87.86 & \underline{71.74} & \textbf{46.61} & 55.51 & 34.40 & \textbf{97.23\%} \\
    TORINO-M & \underline{59.92} & \textbf{66.32} & 60.31 & 1728.88 & 87.94 & 71.64 & \underline{46.53} & \underline{55.56} & 34.40 & \underline{97.04\%} \\
    \midrule
    \rowcolor{gray!15}
    \multicolumn{11}{c}{\textit{Retain $\sim$129 Tokens in Average ($\downarrow$ 78\%)}} \\
    \midrule
    Random & 57.29 & 58.93 & 54.47 & \textbf{1774.34} & 84.50 & 70.60 & 30.06 & 55.67 & 27.80 & 88.26\% \\
    FOLDER & 58.01 & 61.68 & 56.44 & 1693.52 & \underline{86.89} & 71.24 & 35.88 & \underline{56.10} & 30.30 & 91.27\% \\
    PruneSID & \underline{58.43} & 63.75 & 58.93 & 1731.43 & 86.86 & \underline{72.24} & 41.23 & 55.78 & 32.30 & 94.28\% \\
    PruMerge & 57.25 & \textbf{66.07} & 59.36 & 1676.44 & 84.14 & \textbf{72.43} & 42.49 & 55.86 & 33.90 & 94.67\% \\
    TORINO-P & \textbf{58.62} & \underline{65.46} & 59.97 & \underline{1745.80} & \textbf{86.96} & 70.80 & \underline{43.93} & 55.97 & \underline{34.50} & \textbf{95.99\%} \\
    TORINO-M & 58.20 & 65.21 & \underline{59.97} & 1725.12 & 86.50 & 71.10 & \textbf{43.98} & \textbf{56.19} & \textbf{34.60} & \underline{95.82\%} \\
    \midrule
    \rowcolor{gray!15}
    \multicolumn{11}{c}{\textit{Retain $\sim$47 Tokens in Average ($\downarrow$ 92\%)}} \\
    \midrule
    Random & 54.12 & 51.72 & 46.39 & 1640.23 & 77.98 & 65.84 & 21.17 & 56.88 & 28.70 & 81.21\% \\
    FOLDER & 54.68 & 54.21 & 46.82 & 1642.95 & \underline{82.48} & 67.72 & 26.94 & \underline{58.19} & 26.00 & 83.40\% \\
    PruneSID & \textbf{57.03} & 60.40 & 55.67 & \textbf{1727.21} & \textbf{85.98} & 68.57 & \textbf{39.88} & \textbf{58.29} & 30.20 & \textbf{91.74\%} \\
    PruMerge & 53.74 & \underline{61.77} & \underline{56.27} & 1600.72 & 73.89 & \textbf{71.54} & \underline{39.71} & 57.24 & 28.90 & 89.00\% \\
    TORINO-P & \underline{55.76} & \textbf{62.37} & \textbf{56.79} & \underline{1676.53} & 80.74 & \underline{69.51} & 35.66 & 57.43 & \textbf{31.70} & \underline{90.55\%} \\
    TORINO-M & 55.48 & 61.51 & 55.84 & 1666.57 & 80.44 & 69.41 & 35.76 & 57.72 & \underline{31.50} & 90.10\% \\
    \bottomrule
  \end{tabular}}
\end{table*}

\begin{table*}[t]
  \centering
  \caption{\textbf{Fixed budget} performance comparison on LLaVA-1.5-7B across 9 image understanding benchmarks. SAE: Matryoshka BatchTopK 20, $\varepsilon=64$, CLS-only. Best per tier in \textbf{bold}, second-best \underline{underlined}. Relative is the macro-average of per-benchmark ratios against our reproduced baseline.}
  \label{tab:results-llava-v1-5-7b-fb-x64-cls-only}
\resizebox{\textwidth}{!}{%
  \begin{tabular}{lcccccccccc}
    \toprule
    Method & GQA & MMB$^{\text{EN}}$ & MMB$^{\text{CN}}$ & MME & POPE & SQA$^{\text{I}}$ & VQA$^{\text{T}}$ & VizWiz & MMVet & Relative \\
    \midrule
    \rowcolor{gray!15}
    \multicolumn{11}{c}{\textit{Upper Bound, 576 Tokens (100\%)}} \\
    \midrule
    Baseline & 61.93 & 64.00 & 57.90 & 1834.80 & 86.17 & 67.92 & 45.65 & 54.39 & 31.50 & 100.00\% \\
    \midrule
    \rowcolor{gray!15}
    \multicolumn{11}{c}{\textit{Retain $\sim$192 Tokens in Average ($\downarrow$ 67\%)}} \\
    \midrule
    Random & 58.81 & 61.00 & 53.87 & 1742.51 & 84.43 & 66.98 & 33.24 & \underline{55.54} & 28.60 & 93.40\% \\
    FOLDER & 58.78 & 59.79 & 51.72 & \textbf{1744.85} & 85.92 & \textbf{68.42} & 38.96 & 54.72 & 30.60 & 95.14\% \\
    PruneSID & 58.91 & 61.94 & 55.67 & 1730.87 & \underline{86.24} & 67.63 & 40.72 & 54.80 & \underline{31.50} & \underline{96.89\%} \\
    PruMerge & 57.83 & 61.86 & \textbf{56.53} & 1677.53 & 83.81 & 67.67 & 40.67 & \textbf{56.03} & \textbf{32.50} & 96.81\% \\
    $\mathrm{TORINO}_{FB}$-P & \textbf{59.50} & 62.80 & 56.27 & 1706.19 & 85.28 & \underline{67.77} & \underline{41.63} & 54.60 & 30.40 & 96.80\% \\
    $\mathrm{TORINO}_{FB}$-P-r2 & \underline{59.39} & 61.60 & 54.12 & \underline{1743.89} & \textbf{86.36} & 67.08 & 40.87 & 55.19 & 29.80 & 96.14\% \\
    $\mathrm{TORINO}_{FB}$-M & 59.37 & \underline{62.80} & \underline{56.44} & 1733.61 & 84.59 & 67.72 & \textbf{41.63} & 54.09 & 30.80 & \textbf{96.92\%} \\
    $\mathrm{TORINO}_{FB}$-M-r2 & 58.67 & 60.48 & 51.63 & 1706.96 & 85.41 & 66.83 & 40.03 & 54.49 & 28.60 & 94.18\% \\
    \midrule
    \rowcolor{gray!15}
    \multicolumn{11}{c}{\textit{Retain $\sim$128 Tokens in Average ($\downarrow$ 78\%)}} \\
    \midrule
    Random & 57.54 & 57.90 & 49.83 & 1694.13 & 82.52 & 67.13 & 28.75 & \textbf{56.18} & 28.00 & 90.17\% \\
    FOLDER & 56.36 & 56.53 & 47.08 & 1637.30 & 82.57 & 67.18 & 32.13 & 55.91 & 29.40 & 90.12\% \\
    PruneSID & 58.00 & 61.51 & 54.90 & 1683.84 & \underline{85.05} & 67.82 & 39.87 & 56.02 & 30.70 & \underline{95.86\%} \\
    PruMerge & 56.89 & \underline{62.11} & \textbf{55.24} & 1653.06 & 80.98 & 67.92 & 40.00 & \underline{56.12} & \textbf{31.10} & 95.32\% \\
    $\mathrm{TORINO}_{FB}$-P & \textbf{58.53} & 61.77 & \underline{55.07} & \textbf{1755.27} & 83.80 & \underline{68.17} & \textbf{40.23} & 55.84 & 30.70 & \textbf{96.41\%} \\
    $\mathrm{TORINO}_{FB}$-P-r2 & \underline{58.31} & 58.85 & 49.83 & 1651.66 & \textbf{86.64} & 67.58 & 37.06 & 55.00 & 26.90 & 92.21\% \\
    $\mathrm{TORINO}_{FB}$-M & 58.00 & 62.11 & 54.98 & \underline{1717.63} & 83.30 & \textbf{68.32} & \underline{40.13} & 55.49 & 28.90 & 95.36\% \\
    $\mathrm{TORINO}_{FB}$-M-r2 & 56.79 & 54.55 & 44.33 & 1561.81 & 83.18 & 65.89 & 34.83 & 54.67 & 25.30 & 87.70\% \\
    \midrule
    \rowcolor{gray!15}
    \multicolumn{11}{c}{\textit{Retain $\sim$64 Tokens in Average ($\downarrow$ 89\%)}} \\
    \midrule
    Random & 55.08 & 53.87 & 43.56 & 1612.24 & 76.44 & 65.15 & 22.67 & 56.43 & 26.20 & 84.16\% \\
    FOLDER & 52.47 & 49.66 & 37.11 & 1383.88 & 66.08 & 65.20 & 23.73 & 56.52 & 24.60 & 78.72\% \\
    PruneSID & \textbf{57.34} & \textbf{60.14} & \textbf{52.92} & \textbf{1711.81} & \textbf{85.13} & \textbf{67.87} & \underline{38.61} & \underline{56.77} & \textbf{29.00} & \textbf{94.55\%} \\
    PruMerge & 54.37 & \underline{59.36} & \underline{51.46} & 1604.14 & 73.87 & 67.53 & \textbf{39.25} & \textbf{56.86} & 27.40 & \underline{91.06\%} \\
    $\mathrm{TORINO}_{FB}$-P & \underline{55.14} & 59.19 & 50.95 & \underline{1621.95} & 76.70 & \underline{67.58} & 33.10 & 56.42 & 28.20 & 90.24\% \\
    $\mathrm{TORINO}_{FB}$-P-r2 & 54.62 & 50.69 & 41.92 & 1311.90 & \underline{78.07} & 63.36 & 22.08 & 54.90 & 24.30 & 80.18\% \\
    $\mathrm{TORINO}_{FB}$-M & 54.87 & 57.99 & 50.26 & 1610.12 & 74.74 & 66.78 & 32.53 & 56.71 & \underline{28.90} & 89.56\% \\
    $\mathrm{TORINO}_{FB}$-M-r2 & 51.19 & 39.43 & 28.95 & 1135.06 & 66.69 & 63.11 & 18.52 & 54.20 & 19.80 & 69.95\% \\
    \bottomrule
  \end{tabular}}
\end{table*}

\begin{table*}[t]
  \centering
  \caption{LLaVA-1.5-7B, dynamic setting, $\varepsilon=1$. Best per tier in \textbf{bold}, second-best \underline{underlined}. Relative is the macro-average of per-benchmark ratios against our reproduced baseline.}
  \label{tab:results-llava-v1-5-7b-x1-cls-only}
  \resizebox{\textwidth}{!}{%
  \begin{tabular}{lcccccccccc}
    \toprule
    Method & GQA & MMB$^{\text{EN}}$ & MMB$^{\text{CN}}$ & MME & POPE & SQA$^{\text{I}}$ & VQA$^{\text{T}}$ & VizWiz & MMVet & Relative \\
    \midrule
    \rowcolor{gray!15}
    \multicolumn{11}{c}{\textit{Upper Bound, 576 Tokens (100\%)}} \\
    \midrule
    Baseline & 61.93 & 64.00 & 57.90 & 1834.80 & 86.17 & 67.92 & 45.65 & 54.39 & 31.50 & 100.00\% \\
    \midrule
    \rowcolor{gray!15}
    \multicolumn{11}{c}{\textit{Retain $\sim$243 Tokens in Average ($\downarrow$ 58\%)}} \\
    \midrule
    Random & 59.67 & 61.51 & 55.07 & 1759.00 & 84.66 & 67.58 & 35.38 & \underline{55.22} & 30.80 & 95.33\% \\
    FOLDER & 59.79 & 61.25 & 53.44 & 1731.12 & 86.40 & \underline{68.17} & 40.05 & 54.12 & 30.00 & 95.78\% \\
    PruneSID & 59.62 & 62.20 & 56.19 & 1683.08 & 86.23 & 67.97 & 41.19 & 54.47 & 31.00 & 96.79\% \\
    PruMerge & 58.53 & 62.03 & \textbf{56.70} & 1686.31 & 84.65 & 67.48 & 40.51 & \textbf{55.60} & 31.20 & 96.54\% \\
    TORINO-P & \textbf{60.65} & \underline{62.29} & \underline{56.27} & \underline{1777.49} & \underline{86.60} & 67.67 & \textbf{43.78} & 54.56 & \textbf{31.40} & \textbf{98.37\%} \\
    TORINO-M & \underline{60.47} & \textbf{62.37} & 55.58 & \textbf{1803.32} & \textbf{86.71} & \textbf{68.17} & \underline{43.60} & 54.36 & \underline{31.30} & \underline{98.35\%} \\
    \midrule
    \rowcolor{gray!15}
    \multicolumn{11}{c}{\textit{Retain $\sim$123 Tokens in Average ($\downarrow$ 79\%)}} \\
    \midrule
    Random & \underline{57.36} & 57.04 & 35.57 & 1683.51 & 82.25 & 66.83 & 28.27 & 56.04 & 29.10 & 87.35\% \\
    FOLDER & 56.37 & 56.36 & 44.93 & 1610.11 & 82.47 & 66.98 & 32.02 & 55.90 & 30.20 & 89.73\% \\
    PruneSID & 57.18 & \underline{60.31} & \underline{54.47} & \underline{1704.24} & \underline{83.44} & 67.72 & 37.13 & \underline{56.48} & \underline{30.40} & \textbf{94.64\%} \\
    PruMerge & 55.66 & \textbf{61.77} & \textbf{54.90} & 1683.28 & 78.32 & \underline{67.72} & \textbf{38.91} & \textbf{56.52} & \textbf{31.10} & \underline{94.60\%} \\
    TORINO-P & 56.73 & 60.14 & 52.84 & \textbf{1721.81} & \textbf{84.44} & 67.13 & 36.23 & 55.98 & 30.10 & 93.92\% \\
    TORINO-M & \textbf{58.39} & 53.87 & 51.72 & 1694.21 & 73.71 & 66.88 & \underline{37.61} & 55.59 & 28.90 & 91.16\% \\
    \midrule
    \rowcolor{gray!15}
    \multicolumn{11}{c}{\textit{Retain $\sim$37 Tokens in Average ($\downarrow$ 94\%)}} \\
    \midrule
    Random & 52.63 & 48.28 & 35.65 & 1505.69 & 70.55 & 64.90 & 18.67 & 55.99 & 22.60 & 77.45\% \\
    FOLDER & 49.93 & 40.29 & 27.15 & 1210.04 & 54.95 & 66.04 & 19.56 & 54.74 & 23.70 & 70.68\% \\
    PruneSID & \textbf{55.18} & \underline{57.73} & \underline{48.80} & \underline{1590.28} & \textbf{82.21} & \underline{67.18} & \underline{35.93} & \textbf{56.62} & \textbf{26.70} & \textbf{90.24\%} \\
    PruMerge & 51.97 & \textbf{59.19} & \textbf{51.29} & \textbf{1602.39} & 67.94 & \textbf{67.77} & \textbf{37.14} & \underline{56.09} & 23.70 & \underline{87.85\%} \\
    TORINO-P & \underline{53.51} & 54.64 & 44.42 & 1500.45 & \underline{77.37} & 66.88 & 27.16 & 55.60 & \underline{25.90} & 84.72\% \\
    TORINO-M & 53.39 & 51.55 & 39.86 & 1383.84 & 73.50 & 64.60 & 26.70 & 55.27 & 24.40 & 81.00\% \\
    \bottomrule
  \end{tabular}}
\end{table*}

\begin{table*}[t]
  \centering
  \caption{LLaVA-1.5-7B, dynamic setting, $\varepsilon=2$. Best per tier in \textbf{bold}, second-best \underline{underlined}. Relative is the macro-average of per-benchmark ratios against our reproduced baseline.}
  \label{tab:results-llava-v1-5-7b-x2-cls-only}
  \resizebox{\textwidth}{!}{%
  \begin{tabular}{lcccccccccc}
    \toprule
    Method & GQA & MMB$^{\text{EN}}$ & MMB$^{\text{CN}}$ & MME & POPE & SQA$^{\text{I}}$ & VQA$^{\text{T}}$ & VizWiz & MMVet & Relative \\
    \midrule
    \rowcolor{gray!15}
    \multicolumn{11}{c}{\textit{Upper Bound, 576 Tokens (100\%)}} \\
    \midrule
    Baseline & 61.93 & 64.00 & 57.90 & 1834.80 & 86.17 & 67.92 & 45.65 & 54.39 & 31.50 & 100.00\% \\
    \midrule
    \rowcolor{gray!15}
    \multicolumn{11}{c}{\textit{Retain $\sim$290 Tokens in Average ($\downarrow$ 50\%)}} \\
    \midrule
    Random & 60.12 & 61.51 & 55.24 & 1783.74 & 84.88 & 67.87 & 37.40 & 54.69 & 31.30 & 96.23\% \\
    FOLDER & 60.25 & \underline{62.29} & 55.33 & \underline{1810.69} & \textbf{86.86} & \textbf{68.37} & 41.67 & \textbf{55.64} & \textbf{32.70} & 98.63\% \\
    PruneSID & 59.41 & 61.77 & \textbf{56.96} & 1713.47 & \underline{86.35} & \underline{68.02} & 41.30 & 53.96 & 31.20 & 97.03\% \\
    PruMerge & 59.07 & \textbf{63.14} & 56.01 & 1721.68 & 85.79 & 67.28 & 40.29 & \underline{55.20} & 30.50 & 96.64\% \\
    TORINO-P & \textbf{60.98} & 62.20 & \underline{56.10} & 1798.53 & 86.04 & 67.63 & \underline{44.78} & 54.15 & \underline{32.20} & \textbf{98.87\%} \\
    TORINO-M & \underline{60.97} & 62.03 & 55.67 & \textbf{1812.56} & 86.29 & 67.43 & \textbf{44.80} & 53.98 & 32.00 & \underline{98.74\%} \\
    \midrule
    \rowcolor{gray!15}
    \multicolumn{11}{c}{\textit{Retain $\sim$185 Tokens in Average ($\downarrow$ 68\%)}} \\
    \midrule
    Random & 57.89 & 60.82 & 53.26 & \textbf{1777.07} & 83.59 & 67.38 & 30.85 & \textbf{56.54} & 29.10 & 93.05\% \\
    FOLDER & 58.56 & 53.87 & 50.69 & 1682.78 & 85.91 & \textbf{68.57} & 38.70 & 55.03 & 29.20 & 93.03\% \\
    PruneSID & 58.82 & 61.77 & 51.80 & 1731.09 & 85.70 & \underline{68.27} & 38.67 & 55.64 & \underline{31.20} & 95.70\% \\
    PruMerge & 57.48 & \textbf{62.20} & \underline{55.24} & 1713.99 & 83.72 & 67.58 & 40.65 & \underline{56.01} & \textbf{31.30} & \underline{96.32\%} \\
    TORINO-P & \textbf{59.25} & \underline{62.11} & \textbf{55.33} & \underline{1771.73} & \textbf{86.38} & 67.63 & \textbf{41.84} & 55.22 & 30.60 & \textbf{97.22\%} \\
    TORINO-M & \underline{58.86} & 58.68 & 54.30 & 1706.83 & \underline{86.33} & 67.63 & \underline{41.38} & 55.39 & 31.10 & 96.05\% \\
    \midrule
    \rowcolor{gray!15}
    \multicolumn{11}{c}{\textit{Retain $\sim$61 Tokens in Average ($\downarrow$ 89\%)}} \\
    \midrule
    Random & 55.09 & 53.01 & 44.16 & \underline{1611.01} & 76.21 & 66.09 & 22.36 & 56.31 & 26.80 & 84.35\% \\
    FOLDER & 52.35 & 50.26 & 36.68 & 1383.94 & 68.60 & 65.20 & 23.64 & \textbf{56.61} & 24.80 & 79.12\% \\
    PruneSID & \textbf{57.63} & \textbf{59.88} & \textbf{52.06} & \textbf{1720.14} & \textbf{84.43} & 67.13 & \underline{37.98} & 56.02 & \textbf{29.00} & \textbf{93.93\%} \\
    PruMerge & 54.40 & 59.54 & \underline{51.46} & 1601.64 & 74.74 & \textbf{67.63} & \textbf{39.92} & 55.83 & 26.80 & 90.95\% \\
    TORINO-P & \underline{56.57} & \underline{59.71} & 50.09 & 1601.03 & \underline{83.33} & 67.28 & 33.69 & \underline{56.52} & 28.30 & \underline{91.30\%} \\
    TORINO-M & 56.27 & 56.79 & 47.34 & 1536.99 & 81.79 & \underline{67.48} & 33.14 & 56.36 & 28.30 & 89.49\% \\
    \bottomrule
  \end{tabular}}
\end{table*}

\begin{table*}[t]
  \centering
  \caption{LLaVA-1.5-7B, dynamic setting, $\varepsilon=4$. Best per tier in \textbf{bold}, second-best \underline{underlined}. Relative is the macro-average of per-benchmark ratios against our reproduced baseline.}
  \label{tab:results-llava-v1-5-7b-x4-cls-only}
  \resizebox{\textwidth}{!}{%
  \begin{tabular}{lcccccccccc}
    \toprule
    Method & GQA & MMB$^{\text{EN}}$ & MMB$^{\text{CN}}$ & MME & POPE & SQA$^{\text{I}}$ & VQA$^{\text{T}}$ & VizWiz & MMVet & Relative \\
    \midrule
    \rowcolor{gray!15}
    \multicolumn{11}{c}{\textit{Upper Bound, 576 Tokens (100\%)}} \\
    \midrule
    Baseline & 61.93 & 64.00 & 57.90 & 1834.80 & 86.17 & 67.92 & 45.65 & 54.39 & 31.50 & 100.00\% \\
    \midrule
    \rowcolor{gray!15}
    \multicolumn{11}{c}{\textit{Retain $\sim$365 Tokens in Average ($\downarrow$ 37\%)}} \\
    \midrule
    Random & 60.67 & 63.14 & 56.53 & 1764.15 & 85.57 & 67.58 & 40.51 & 54.89 & 30.80 & 97.40\% \\
    FOLDER & \underline{61.35} & \underline{63.57} & 57.13 & \textbf{1778.81} & \textbf{86.34} & \textbf{67.97} & 44.27 & \underline{55.42} & 31.00 & 99.06\% \\
    PruneSID & 59.73 & 62.20 & \underline{57.30} & 1743.41 & \underline{86.29} & \underline{67.67} & 41.49 & 53.62 & 30.80 & 97.18\% \\
    PruMerge & 59.25 & 62.89 & 55.67 & 1711.74 & 86.14 & 67.23 & 40.43 & \textbf{55.55} & 29.40 & 96.26\% \\
    TORINO-P & 61.32 & 63.32 & 57.30 & 1767.33 & 85.63 & 67.53 & \textbf{44.88} & 54.07 & \underline{32.90} & \underline{99.35\%} \\
    TORINO-M & \textbf{61.44} & \textbf{63.66} & 57.04 & \underline{1771.54} & 85.58 & 67.28 & \underline{44.73} & 54.20 & \textbf{33.10} & \textbf{99.43\%} \\
    \midrule
    \rowcolor{gray!15}
    \multicolumn{11}{c}{\textit{Retain $\sim$284 Tokens in Average ($\downarrow$ 51\%)}} \\
    \midrule
    Random & 59.59 & 61.34 & 55.33 & \textbf{1823.68} & 85.27 & 67.43 & 34.09 & \textbf{55.18} & 31.30 & 95.63\% \\
    FOLDER & 59.30 & 61.77 & 54.12 & 1721.99 & \textbf{86.20} & \textbf{68.67} & 37.90 & 54.85 & 29.90 & 95.50\% \\
    PruneSID & 59.45 & 61.68 & \underline{56.36} & 1704.11 & \underline{86.14} & \underline{68.17} & \textbf{41.64} & 53.83 & 31.40 & 96.98\% \\
    PruMerge & 58.12 & \textbf{62.54} & 55.41 & 1698.90 & 82.36 & 67.48 & 40.47 & \underline{55.12} & 30.00 & 95.56\% \\
    TORINO-P & \textbf{60.82} & \underline{62.29} & \textbf{56.87} & \underline{1800.80} & 86.12 & 67.18 & \underline{41.17} & 54.44 & \underline{33.40} & \textbf{98.56\%} \\
    TORINO-M & \underline{60.60} & 62.20 & 56.01 & 1769.19 & 86.14 & 67.58 & 41.04 & 54.46 & \textbf{33.50} & \underline{98.23\%} \\
    \midrule
    \rowcolor{gray!15}
    \multicolumn{11}{c}{\textit{Retain $\sim$109 Tokens in Average ($\downarrow$ 81\%)}} \\
    \midrule
    Random & 56.84 & 56.27 & 49.05 & 1688.46 & 81.37 & 66.34 & 27.09 & 56.02 & 28.20 & 88.93\% \\
    FOLDER & 55.97 & 55.76 & 45.88 & 1604.52 & 81.45 & 66.78 & 31.53 & \textbf{56.63} & 26.80 & 88.37\% \\
    PruneSID & 57.61 & \underline{60.74} & \underline{54.21} & \underline{1711.23} & 84.28 & \underline{67.58} & \underline{39.07} & 55.88 & 30.00 & \underline{95.07\%} \\
    PruMerge & 56.28 & \textbf{62.11} & \textbf{54.47} & 1684.78 & 79.76 & \textbf{67.87} & \textbf{40.23} & \underline{56.51} & \underline{30.60} & 95.05\% \\
    TORINO-P & \textbf{58.41} & 60.57 & 53.52 & \textbf{1728.62} & \textbf{85.71} & 66.68 & 37.81 & 56.02 & \textbf{30.70} & \textbf{95.17\%} \\
    TORINO-M & \underline{57.97} & 58.93 & 50.34 & 1682.49 & \underline{84.30} & 67.28 & 36.38 & 55.90 & 28.00 & 92.51\% \\
    \bottomrule
  \end{tabular}}
\end{table*}

\begin{table*}[t]
  \centering
  \caption{LLaVA-1.5-7B, dynamic setting, $\varepsilon=8$. Best per tier in \textbf{bold}, second-best \underline{underlined}. Relative is the macro-average of per-benchmark ratios against our reproduced baseline.}
  \label{tab:results-llava-v1-5-7b-x8-cls-only}
  \resizebox{\textwidth}{!}{%
  \begin{tabular}{lcccccccccc}
    \toprule
    Method & GQA & MMB$^{\text{EN}}$ & MMB$^{\text{CN}}$ & MME & POPE & SQA$^{\text{I}}$ & VQA$^{\text{T}}$ & VizWiz & MMVet & Relative \\
    \midrule
    \rowcolor{gray!15}
    \multicolumn{11}{c}{\textit{Upper Bound, 576 Tokens (100\%)}} \\
    \midrule
    Baseline & 61.93 & 64.00 & 57.90 & 1834.80 & 86.17 & 67.92 & 45.65 & 54.39 & 31.50 & 100.00\% \\
    \midrule
    \rowcolor{gray!15}
    \multicolumn{11}{c}{\textit{Retain $\sim$312 Tokens in Average ($\downarrow$ 46\%)}} \\
    \midrule
    Random & 60.43 & 62.03 & 55.76 & 1797.24 & 85.13 & \underline{67.97} & 38.34 & 54.89 & 29.70 & 96.31\% \\
    FOLDER & 60.86 & \underline{62.71} & 56.19 & \underline{1809.73} & \textbf{86.50} & \textbf{68.02} & 43.40 & \underline{55.42} & 30.20 & 98.37\% \\
    PruneSID & 59.36 & 62.37 & 56.19 & 1745.44 & \underline{86.10} & 67.72 & 41.52 & 54.20 & 31.30 & 97.23\% \\
    PruMerge & 59.06 & \textbf{62.89} & 56.01 & 1718.73 & 85.75 & 67.33 & 40.66 & \textbf{55.66} & 31.30 & 97.05\% \\
    TORINO-P & \underline{61.20} & 62.29 & \underline{56.70} & 1790.78 & 85.76 & 67.38 & \underline{44.51} & 54.56 & \textbf{33.60} & \textbf{99.43\%} \\
    TORINO-M & \textbf{61.27} & 62.54 & \textbf{56.87} & \textbf{1828.29} & 85.91 & 67.03 & \textbf{44.96} & 54.17 & \underline{31.90} & \underline{99.14\%} \\
    \midrule
    \rowcolor{gray!15}
    \multicolumn{11}{c}{\textit{Retain $\sim$150 Tokens in Average ($\downarrow$ 74\%)}} \\
    \midrule
    Random & \underline{58.47} & 50.95 & 51.63 & 1666.99 & 82.01 & \underline{67.87} & 30.29 & \textbf{56.68} & 31.20 & 90.97\% \\
    FOLDER & 50.82 & 44.85 & 48.28 & 1658.45 & \underline{84.17} & 67.38 & 35.19 & \underline{56.37} & 30.20 & 88.82\% \\
    PruneSID & 57.88 & \textbf{62.03} & \underline{54.47} & \underline{1762.75} & 83.89 & 67.28 & 37.42 & 56.14 & 31.20 & \underline{95.68\%} \\
    PruMerge & 57.29 & 58.93 & \textbf{55.41} & 1697.74 & 82.02 & \textbf{68.02} & \underline{40.50} & 56.15 & \textbf{31.60} & 95.60\% \\
    TORINO-P & \textbf{59.02} & \underline{61.25} & 54.12 & \textbf{1823.11} & \textbf{85.50} & 66.98 & \textbf{40.94} & 55.44 & \underline{31.40} & \textbf{97.00\%} \\
    TORINO-M & 56.02 & 61.17 & 53.09 & 1735.12 & 82.84 & 67.03 & 39.03 & 56.13 & 31.30 & 95.02\% \\
    \midrule
    \rowcolor{gray!15}
    \multicolumn{11}{c}{\textit{Retain $\sim$38 Tokens in Average ($\downarrow$ 93\%)}} \\
    \midrule
    Random & \underline{52.89} & 48.28 & 35.40 & 1518.59 & 71.61 & 65.34 & 18.43 & 56.08 & \underline{26.70} & 79.14\% \\
    FOLDER & 49.84 & 40.98 & 28.44 & 1212.22 & 47.05 & \underline{66.83} & 20.36 & 55.02 & 24.30 & 70.62\% \\
    PruneSID & \textbf{55.26} & \underline{57.73} & \underline{49.14} & \underline{1592.23} & \textbf{82.56} & 66.04 & \underline{35.91} & \textbf{56.54} & \textbf{30.80} & \textbf{91.61\%} \\
    PruMerge & 52.42 & \textbf{59.45} & \textbf{50.60} & \textbf{1642.63} & 68.37 & \textbf{67.33} & \textbf{37.09} & 56.28 & 26.50 & \underline{89.09\%} \\
    TORINO-P & 52.88 & 55.67 & 46.82 & 1464.32 & \underline{73.02} & 64.40 & 31.24 & \underline{56.40} & 24.00 & 84.55\% \\
    TORINO-M & 52.05 & 54.30 & 44.76 & 1457.02 & 71.21 & 64.35 & 30.10 & 55.86 & 23.40 & 82.88\% \\
    \bottomrule
  \end{tabular}}
\end{table*}

\begin{table*}[t]
  \centering
  \caption{LLaVA-1.5-7B, dynamic setting, $\varepsilon=16$. Best per tier in \textbf{bold}, second-best \underline{underlined}. Relative is the macro-average of per-benchmark ratios against our reproduced baseline.}
  \label{tab:results-llava-v1-5-7b-x16-cls-only}
  \resizebox{\textwidth}{!}{%
  \begin{tabular}{lcccccccccc}
    \toprule
    Method & GQA & MMB$^{\text{EN}}$ & MMB$^{\text{CN}}$ & MME & POPE & SQA$^{\text{I}}$ & VQA$^{\text{T}}$ & VizWiz & MMVet & Relative \\
    \midrule
    \rowcolor{gray!15}
    \multicolumn{11}{c}{\textit{Upper Bound, 576 Tokens (100\%)}} \\
    \midrule
    Baseline & 61.93 & 64.00 & 57.90 & 1834.80 & 86.17 & 67.92 & 45.65 & 54.39 & 31.50 & 100.00\% \\
    \midrule
    \rowcolor{gray!15}
    \multicolumn{11}{c}{\textit{Retain $\sim$305 Tokens in Average ($\downarrow$ 47\%)}} \\
    \midrule
    Random & 60.27 & 62.20 & 55.41 & 1775.33 & 84.93 & 67.72 & 38.01 & 54.98 & 30.50 & 96.27\% \\
    FOLDER & 60.78 & 61.94 & \underline{56.19} & \textbf{1810.09} & \underline{86.44} & \textbf{67.82} & 43.32 & \underline{55.70} & \underline{32.00} & \underline{98.85\%} \\
    PruneSID & 59.59 & 62.11 & \textbf{56.27} & 1742.91 & \textbf{86.55} & \underline{67.77} & 41.48 & 53.98 & 31.30 & 97.24\% \\
    PruMerge & 59.10 & \underline{62.46} & 55.93 & 1735.85 & 85.57 & 66.83 & 40.55 & \textbf{55.74} & 30.40 & 96.64\% \\
    TORINO-P & \underline{61.07} & 61.77 & 55.93 & \underline{1802.93} & 86.19 & 67.63 & \underline{44.73} & 54.28 & 31.40 & 98.56\% \\
    TORINO-M & \textbf{61.18} & 62.46 & 56.01 & 1796.98 & 86.22 & 67.67 & \textbf{44.94} & 54.22 & \textbf{33.10} & \textbf{99.33\%} \\
    \midrule
    \rowcolor{gray!15}
    \multicolumn{11}{c}{\textit{Retain $\sim$136 Tokens in Average ($\downarrow$ 76\%)}} \\
    \midrule
    Random & 55.00 & 37.80 & 49.14 & 1628.24 & 81.04 & 67.53 & 27.15 & \textbf{56.93} & 29.00 & 85.68\% \\
    FOLDER & 54.44 & 56.36 & 45.53 & 1589.60 & 80.47 & 67.53 & 30.87 & 56.06 & 29.90 & 88.85\% \\
    PruneSID & \textbf{56.89} & \underline{61.25} & \underline{55.07} & \textbf{1752.13} & \textbf{81.62} & 67.67 & 34.62 & \underline{56.49} & 29.70 & \textbf{94.06\%} \\
    PruMerge & 54.45 & \textbf{62.29} & \textbf{55.50} & 1268.76 & 79.71 & \textbf{68.12} & 30.78 & 56.31 & 31.20 & 90.34\% \\
    TORINO-P & 54.33 & 60.91 & 53.78 & 1724.19 & \underline{81.57} & \underline{68.02} & \underline{35.37} & 56.42 & 31.20 & \underline{93.87\%} \\
    TORINO-M & \underline{56.77} & 50.86 & 52.84 & \underline{1745.11} & 65.11 & 67.58 & \textbf{37.06} & 55.80 & \textbf{33.20} & 91.30\% \\
    \midrule
    \rowcolor{gray!15}
    \multicolumn{11}{c}{\textit{Retain $\sim$17 Tokens in Average ($\downarrow$ 97\%)}} \\
    \midrule
    Random & \underline{48.73} & 37.37 & 21.99 & \underline{1334.50} & \underline{54.71} & 63.61 & 14.58 & 54.55 & 16.20 & 65.40\% \\
    FOLDER & 46.22 & 26.63 & 14.35 & 955.87 & 16.84 & 62.32 & 14.77 & 53.36 & 20.20 & 55.45\% \\
    PruneSID & \textbf{50.99} & \textbf{52.84} & \textbf{42.70} & \textbf{1488.90} & \textbf{66.42} & \underline{67.23} & \underline{30.92} & \textbf{55.30} & \textbf{24.50} & \textbf{82.56\%} \\
    PruMerge & 47.32 & \underline{50.09} & \underline{39.86} & 1237.04 & 54.42 & \textbf{68.42} & \textbf{32.38} & \underline{55.01} & \underline{24.10} & \underline{78.15\%} \\
    TORINO-P & 47.59 & 47.16 & 36.86 & 1170.37 & 51.12 & 63.81 & 25.41 & 54.37 & 21.20 & 72.69\% \\
    TORINO-M & 47.16 & 44.16 & 34.71 & 1180.13 & 49.18 & 63.66 & 24.84 & 53.76 & 21.20 & 71.20\% \\
    \bottomrule
  \end{tabular}}
\end{table*}

\begin{table*}[t]
  \centering
  \caption{Performance comparison on LLaVA-1.5-7B across 9 image understanding benchmarks. SAE: Matryoshka BatchTopK 20 $\varepsilon 64$, trained on \textbf{two random spatial tokens per image}. Best per tier in \textbf{bold}, second-best \underline{underlined}. Relative is the macro-average of per-benchmark ratios against our reproduced baseline.}
  \label{tab:results-llava-v1-5-7b-x64-random-k-2}
  \resizebox{\textwidth}{!}{%
  \begin{tabular}{lcccccccccc}
    \toprule
    Method & GQA & MMB$^{\text{EN}}$ & MMB$^{\text{CN}}$ & MME & POPE & SQA$^{\text{I}}$ & VQA$^{\text{T}}$ & VizWiz & MMVet & Relative \\
    \midrule
    \rowcolor{gray!15}
    \multicolumn{11}{c}{\textit{Upper Bound, 576 Tokens (100\%)}} \\
    \midrule
    Baseline & 61.93 & 64.00 & 57.90 & 1834.80 & 86.17 & 67.92 & 45.65 & 54.39 & 31.50 & 100.00\% \\
    \midrule
    \rowcolor{gray!15}
    \multicolumn{11}{c}{\textit{Retain $\sim$205 Tokens in Average ($\downarrow$ 64\%)}} \\
    \midrule
    Random & 58.90 & 61.17 & 54.30 & 1752.00 & 84.48 & 67.23 & 33.60 & \underline{55.55} & 29.90 & 94.18\% \\
    FOLDER & 59.11 & 60.05 & 52.15 & 1757.36 & 85.91 & \textbf{68.27} & 39.29 & 54.34 & \underline{30.30} & 95.28\% \\
    PruneSID & 58.88 & 62.11 & \underline{55.67} & \underline{1761.29} & 85.84 & \underline{68.17} & 40.94 & 54.40 & 29.30 & 96.33\% \\
    PruMerge & 57.96 & \underline{62.11} & \textbf{56.36} & 1700.25 & 84.53 & 67.67 & 40.85 & \textbf{55.79} & \textbf{30.80} & \underline{96.47\%} \\
    TORINO-P-r2 & \textbf{60.30} & 61.17 & 53.61 & \textbf{1795.89} & \textbf{87.12} & 67.28 & \textbf{42.54} & 54.63 & 28.90 & \textbf{96.55\%} \\
    TORINO-M-r2 & \underline{59.65} & 59.62 & 50.95 & 1754.94 & \underline{86.17} & 66.68 & \underline{41.66} & 54.35 & 28.40 & 94.74\% \\
    \midrule
    \rowcolor{gray!15}
    \multicolumn{11}{c}{\textit{Retain $\sim$108 Tokens in Average ($\downarrow$ 81\%)}} \\
    \midrule
    Random & 51.84 & 58.08 & 47.94 & 1681.06 & 81.17 & 67.48 & 25.60 & \underline{56.70} & 27.30 & 87.71\% \\
    FOLDER & 55.78 & 55.67 & 45.96 & 1593.14 & 81.03 & 67.03 & 31.46 & \textbf{56.72} & 26.60 & 88.18\% \\
    PruneSID & \textbf{58.09} & \underline{60.57} & \underline{54.21} & \textbf{1699.84} & \underline{84.35} & \underline{67.82} & \underline{39.07} & 55.91 & \underline{30.20} & \textbf{95.19\%} \\
    PruMerge & 56.34 & \textbf{61.86} & \textbf{55.07} & \underline{1684.29} & 79.43 & \textbf{68.12} & \textbf{40.18} & 56.58 & \textbf{30.60} & \underline{95.13\%} \\
    TORINO-P-r2 & \underline{57.14} & 58.42 & 49.40 & 1622.17 & \textbf{85.46} & 67.08 & 35.53 & 55.32 & 26.70 & 91.06\% \\
    TORINO-M-r2 & 54.90 & 54.55 & 43.56 & 1425.71 & 81.95 & 65.54 & 33.02 & 54.74 & 26.20 & 86.06\% \\
    \midrule
    \rowcolor{gray!15}
    \multicolumn{11}{c}{\textit{Retain $\sim$34 Tokens in Average ($\downarrow$ 94\%)}} \\
    \midrule
    Random & 52.44 & 55.84 & 33.85 & 1466.84 & 67.46 & 65.39 & 18.08 & 55.89 & \underline{25.10} & 78.55\% \\
    FOLDER & 49.05 & 39.52 & 26.37 & 1178.72 & 42.10 & 63.96 & 18.39 & 54.67 & 23.50 & 67.68\% \\
    PruneSID & \textbf{54.77} & \underline{57.73} & \underline{48.71} & \textbf{1607.40} & \textbf{80.85} & \underline{66.53} & \textbf{37.63} & \textbf{56.96} & 24.20 & \textbf{89.57\%} \\
    PruMerge & 51.69 & \textbf{58.59} & \textbf{50.00} & \underline{1606.64} & 67.38 & \textbf{67.48} & \underline{33.20} & \underline{56.18} & \textbf{26.90} & \underline{87.54\%} \\
    TORINO-P-r2 & \underline{54.59} & 50.77 & 40.89 & 1324.22 & \underline{78.04} & 62.82 & 22.65 & 55.11 & 23.80 & 79.98\% \\
    TORINO-M-r2 & 51.38 & 40.21 & 27.58 & 1152.58 & 68.05 & 62.57 & 19.46 & 54.25 & 19.70 & 70.25\% \\
    \bottomrule
  \end{tabular}}
\end{table*}

\begin{table*}[h]
  \centering
  \caption{\textbf{Performance comparison on LLaVA-NeXT-7B across 9 image understanding benchmarks}. SAE: Matryoshka BatchTopK 20 $\varepsilon = 64$, CLS-only. Best per tier in \textbf{bold}, second-best \underline{underlined}. Relative is the macro-average of per-benchmark ratios against our reproduced baseline.}
  \label{tab:results-llava-v1-6-7b-x64-cls-only}
  \resizebox{\textwidth}{!}{%
  \begin{tabular}{lcccccccccc}
\toprule
    Method & GQA & MMB$^{\text{EN}}$ & MMB$^{\text{CN}}$ & MME & POPE & SQA$^{\text{I}}$ & VQA$^{\text{T}}$ & VizWiz & MMVet & Relative \\
    \midrule
    \rowcolor{gray!15}
    \multicolumn{11}{c}{\textit{Upper Bound, 2117 Tokens (100\%)}} \\
    \midrule
    Baseline & 64.2 & 64.5 & 54.7 & 1643.1 & 87.2 & 63.8 & 64.5 & 60.8 & 40.9 & 100.0\% \\
    \midrule
    \rowcolor{gray!15}
    \multicolumn{11}{c}{\textit{Retain $\sim$772 Tokens in Average ($\downarrow$ 63\%)}} \\
    \midrule
    PruneSID & \underline{63.5} & \underline{64.5} & \underline{55.7} & 1661.6 & \textbf{87.5} & 63.3 & 60.1 & \textbf{60.5} & 37.6 & 98.4\% \\
    VisionZip & 63.2 & 63.7 & \textbf{55.8} & 1666.3 & \underline{87.0} & \underline{63.8} & \textbf{63.0} & \underline{60.3} & 38.9 & \underline{99.2\%} \\
    TORINO-P & \textbf{63.5} & 64.3 & 55.6 & \textbf{1691.5} & 86.2 & 63.7 & \underline{61.1} & 60.2 & 40.3 & \textbf{99.4\%} \\
    TORINO-M & 63.1 & \textbf{64.7} & 55.6 & \underline{1673.2} & 86.2 & \textbf{63.8} & 60.6 & 59.6 & \underline{40.3} & 99.1\% \\
    \midrule
    \rowcolor{gray!15}
    \multicolumn{11}{c}{\textit{Retain $\sim$429 Tokens in Average ($\downarrow$ 80\%)}} \\
    \midrule
    PruneSID & \textbf{61.9} & 63.3 & 53.4 & 1642.6 & \textbf{86.6} & 63.6 & 56.6 & \underline{58.5} & 34.8 & 95.6\% \\
    VisionZip & 60.6 & \textbf{63.7} & 54.3 & 1655.5 & \underline{85.2} & \textbf{64.3} & \textbf{60.5} & \textbf{58.7} & 35.7 & \underline{96.6\%} \\
    TORINO-P & \underline{61.4} & 63.5 & \underline{54.5} & \textbf{1718.8} & 84.4 & \underline{64.0} & \underline{57.1} & 58.1 & \textbf{37.0} & \textbf{96.7\%} \\
    TORINO-M & 61.0 & \underline{63.6} & \textbf{54.6} & \underline{1677.4} & 83.3 & 63.0 & 56.3 & 56.9 & \underline{36.8} & 95.6\% \\
    \midrule
    \rowcolor{gray!15}
    \multicolumn{11}{c}{\textit{Retain $\sim$163 Tokens in Average ($\downarrow$ 92\%)}} \\
    \midrule
    PruneSID & \textbf{57.7} & \textbf{60.5} & \underline{48.4} & \underline{1525.6} & \textbf{82.3} & \underline{64.4} & \underline{48.1} & \underline{55.4} & \underline{29.7} & \textbf{88.7\%} \\
    VisionZip & 55.2 & 59.6 & 48.2 & 1484.6 & \underline{74.8} & \textbf{65.2} & \textbf{53.9} & \textbf{55.5} & 29.6 & \underline{88.0\%} \\
    TORINO-P & \underline{56.8} & \underline{59.8} & \textbf{49.5} & \textbf{1540.0} & 74.0 & 63.1 & 46.2 & 54.9 & \textbf{30.4} & 87.3\% \\
    TORINO-M & 55.3 & 57.4 & 45.4 & 1474.0 & 70.7 & 62.6 & 44.6 & 53.3 & 28.4 & 83.7\% \\
    \bottomrule
  \end{tabular}}
\end{table*}

\begin{table*}
  \centering
  \caption{\textbf{Comparison of five seeds with Random on LLaVA-1.5-7B across 9 image understanding benchmarks.} SAE: Matryoshka BatchTopK 20 $\varepsilon = 64$, CLS-only. Best per tier in \textbf{bold}.}
  \label{tab:results-llava-v1-5-7b-x64-cls-only-randomseeds}
  \resizebox{\textwidth}{!}{%
  \begin{tabular}{lcccccccccc}
        \toprule
    Method & GQA & MMB$^{\text{EN}}$ & MMB$^{\text{CN}}$ & MME & POPE & SQA$^{\text{I}}$ & VQA$^{\text{T}}$ & VizWiz & MMVet & Relative \\
    \midrule
    \rowcolor{gray!15}
    \multicolumn{11}{c}{\textit{Upper Bound, 576 Tokens (100\%)}} \\
    \midrule
    Baseline & 61.93 & 64.00 & 57.90 & 1834.80 & 86.17 & 67.92 & 45.65 & 54.39 & 31.50 & 100.00\% \\
    \midrule
    \rowcolor{gray!15}
    \multicolumn{11}{c}{\textit{Retain $\sim$217 Tokens in Average ($\downarrow$ 62\%)}} \\
    \midrule
    Random & 59.45 & 60.48 & 54.73 & 1773.78 & 84.54 & 66.68 & 34.61 & \textbf{55.96} & 28.30 & 94.06\% \\
    Random & 59.06 & 61.17 & 54.12 & 1772.62 & 84.72 & 67.67 & 34.33 & 55.43 & 29.80 & 94.52\% \\
    Random & 59.64 & 61.08 & 53.44 & 1736.95 & 84.87 & \textbf{68.37} & 34.55 & 55.33 & 30.60 & 94.71\% \\
    Random & 59.60 & 60.57 & 53.61 & 1742.84 & 84.24 & 67.28 & 34.40 & 55.79 & 28.90 & 93.88\% \\
    Random & 59.35 & 61.34 & 54.38 & 1770.34 & 84.99 & 66.78 & 34.13 & 55.83 & 30.70 & 94.88\% \\
    TORINO-M & \textbf{60.28} & \textbf{62.20} & \textbf{56.36} & \textbf{1778.78} & 85.30 & 67.58 & \textbf{43.63} & 54.61 & \textbf{33.10} & \textbf{98.70\%} \\
    \bottomrule
  \end{tabular}}
\end{table*}

\end{document}